\def\year{2022}\relax
\documentclass[letterpaper]{article} 
\usepackage{aaai22}  
\usepackage{times}  
\usepackage{helvet}  
\usepackage{courier}  
\usepackage[hyphens]{url}  
\usepackage{graphicx} 
\usepackage{natbib}  
\usepackage{caption} 
\DeclareCaptionStyle{ruled}{labelfont=normalfont,labelsep=colon,strut=off} 
\usepackage{url}            
\usepackage{booktabs}       
\usepackage{amsfonts}       
\usepackage{nicefrac}       
\usepackage{microtype}      
\usepackage{xcolor}         
\usepackage{graphicx}
\usepackage{multirow}
\usepackage{amsmath}
\usepackage{algorithm}  
\usepackage{algpseudocode}  
\usepackage{amsmath}  
\usepackage{amssymb}
\usepackage{xcolor}
\usepackage{listings}
\usepackage{pdfpages}
\definecolor{codegreen}{rgb}{0,0.6,0}
\definecolor{codegray}{rgb}{0.5,0.5,0.5}
\definecolor{codepurple}{rgb}{0.58,0,0.82}
\definecolor{backcolour}{rgb}{0.95,0.95,0.92}

\lstdefinestyle{mystyle}{
  commentstyle=\color{codegreen},
  keywordstyle=\color{magenta},
  numberstyle=\tiny\color{codegray},
  stringstyle=\color{codepurple},
  basicstyle=\ttfamily\footnotesize,
  breakatwhitespace=false,         
  breaklines=true,                 
  captionpos=b,                    
  keepspaces=true,
  numbersep=0pt,                  
  showspaces=false,                
  showstringspaces=false,
  showtabs=false,                  
  tabsize=1,
  numbers=none,
  xleftmargin=0em,
  xrightmargin=0em,
  linewidth=1.0\linewidth,
  aboveskip=0pt,
  belowskip=0pt
}

\title{Focus Your Distribution: Coarse-to-Fine Non-Contrastive Learning for Anomaly Detection and Localization}
\author{
    Ye Zheng\textsuperscript{\rm 2,3},
    Xiang Wang\textsuperscript{\rm 1},
    Rui Deng\textsuperscript{\rm 4},
    Tianpeng Bao\textsuperscript{\rm 1},
    Rui Zhao\textsuperscript{\rm 1},
    Liwei Wu\textsuperscript{\rm 1}
}
\affiliations{
    \textsuperscript{\rm 1} SenseTime Research\\
    \textsuperscript{\rm 2}Institute of Computing Technology, Chinese Academy of Sciences\\
    \textsuperscript{\rm 3}University of Chinese Academy of Sciences
    \textsuperscript{\rm 4}University of California, Los Angeles\\
     
}

\begin{document}

\maketitle

\begin{abstract}
The essence of unsupervised anomaly detection is to learn the compact distribution of normal samples and detect outliers as anomalies in testing. Meanwhile, the anomalies in real-world are usually subtle and fine-grained in a high-resolution image especially for industrial applications. Towards this end, we propose a novel framework for unsupervised anomaly detection and localization. Our method aims at learning dense and compact distribution from normal images with a \textbf{coarse-to-fine alignment} process. The coarse alignment stage standardizes the pixel-wise position of objects in both image and feature levels. The fine alignment stage then densely maximizes the similarity of features among all corresponding locations in a batch. To facilitate the learning with only normal images, we propose a new pretext task called \textbf{non-contrastive learning} for the fine alignment stage. Non-contrastive learning extracts robust and discriminating normal image representations without making assumptions on abnormal samples, and it thus empowers our model to generalize to various anomalous scenarios. Extensive experiments on two typical industrial datasets of MVTec AD and BenTech AD demonstrate that our framework is effective in detecting various real-world defects and achieves a new state-of-the-art in industrial unsupervised anomaly detection.
\end{abstract}
\section{Introduction}
\label{sec:intro}
Image anomaly detection is the identification of unexpected or abnormal image patterns in the dataset, which has wide applications in spotting defects from manufacturing lines~\cite{bergmann2019mvtec}, analyzing medical images~\cite{seebock2016identifying}, and monitoring video streams~\cite{sultani2018real, liu2018classifier}. Different from classical supervised learning tasks that assume an even distribution among classes, the anomalies occur rarely in real-world and are often hard to collect and label. Moreover, the lack of prior knowledge about anomalous patterns imposes a great challenge for designing comprehensive anomaly detection algorithms.

Due to the scarcity and uncertainty of abnormal images, existing anomaly detection methods usually follow the \emph{unsupervised} or \emph{one-class classification} setting. That is, models are provided with only normal data in training. During inference, the anomaly is spotted by the difference between the test data and learned normal features.~\cite{tax1999support,tax2004support,scholkopf1999support}. Existing works~\cite{scholkopf1999support,masci2011stacked,golan2018deep,ruff2018deep,hendrycks2019using} are proven to be successful in abstracting semantically rich representation for isolating defect images; nonetheless, they lack the ability to explore the fine-grained structures for anomalies. For example, a common setting in previous works~\cite{golan2018deep,sohn2020learning} is to set one category in CIFAR-10 dataset~\cite{krizhevsky2009learning} as the normal class and the rest as anomalies. In actual manufacturing or medical industries, however, the difference between normal images and anomalies is more fine-grained and subtle than these object class differences~\cite{bergmann2019mvtec}.

We thus designed a novel framework targeting the fine-grained anomalous patterns in actual industrial setting, where images are usually taken under a clean background and shared positions and defects are usually subtle. The intuition of our method is inspired by the human inference process. When asked to play the \emph{spot the difference} game, human beings would usually first roughly align, or find the correspondence, between the global context of two images. Then, they closely examine the detailed local distinction underlying two patterns. Inspirited by this, we design a two-stage coarse-to-fine framework that learns robust feature distributions for normal images.

We first apply a coarse alignment module to roughly extract and align global feature embeddings. The module operates on both pixel-level for the input image and feature-level for each pyramid feature map. In the fine alignment stage, we apply self-supervised learning and propose a novel pretext task for learning the normal representation. The current state-of-the-art~\cite{li2021cutpaste} in self-supervised anomaly detection designs augmentations that generate abnormal samples through mixing normal image patches. However, we lack sufficient prior knowledge of the real-world anomaly distributions, so the created defects cannot model the numerous real-life possibilities of anomalies. We thus define a new task for self-supervised learning called \textbf{non-contrastive learning} - using no abnormal samples and only normal images to train a robust feature encoder. By enforcing the similarity among each position's feature from a minibatch, we capture the local fine-grained correspondence in every aligned position of images. The distribution of normal images thus becomes more compact and more semantically meaningful, making the abnormal outliers more salient and easier to detect.

To summarize, the main contributions of this paper are:
\begin{itemize}
\item We propose a coarse-to-fine anomaly detection paradigm to detect and localize the fine-grained defects in real-world industrial dataset;
\item We propose a novel pretext task named dense non-contrastive learning for self-supervised learning of compact normal features without any assumption of abnormal samples;
\item We provide extensive experimental results and ablation studies to highlight the strength of our method, and the results in MvTec anomaly detection dataset~\cite{bergmann2019mvtec} show that our method outperforms the previous state-of-the-art anomaly detection methods.
\end{itemize}

\section{Related Work}
The mainstream unsupervised anomaly detection and localization methods are either reconstruction-based or representation-based. 

\textbf{Reconstruction-based method} applies autoencoders~\cite{bergmann2019mvtec, gong2019memorizing} or generative adversarial networks~\cite{sabokrou2018adversarially, pidhorskyi2018generative} to encode and reconstruct the normal data. During inference, an anomaly is spotted when the reconstructed image diverges from the original one. The pixel-wise reconstruction error can be applied to localize anomalies~\cite{bergmann2019mvtec}, and the image level anomaly score is thus determined by aggregating pixel-wise errors~\cite{gong2019memorizing}. Despite the high interpretability of reconstruction and comparison, the pixel-wise difference fails to encode the global semantic meaning of images~\cite{ren2019likelihood, li2021cutpaste}. Also, the autoencoder sometimes generates comparable reconstruction results for the anomalous images too~\cite{perera2019ocgan}. 

\textbf{Representation-based method}, on the other hand, extracts discriminative feature vectors from normal images~\cite{ruff2018deep, bergman2020classification, rippel2021modeling} or normal image patches\cite{bergmann2020uninformed, yi2020patch, cohen2020sub} and yields more promising results for anomaly detection. The anomaly score is calculated by the distance between the embedding of a test image and the distribution of normal image representations. The normal image distribution is typically characterized by the center of a n-sphere for the normal image~\cite{ruff2018deep, yi2020patch}, the Gaussian distribution of normal images~\cite{rippel2021modeling, defard2020padim}, or the kNN for the entire normal image embedding~\cite{bergman2020deep, cohen2020sub}. One of the most recent works, PaDiM~\cite{defard2020padim}, learns the parameters of multivariate Gaussian distribution from different CNN layers. As a concurrent work with ours, PANDAS~\cite{reiss2021panda} also uses non-contrastive learning to achieve feature adaptation to further refine the pretrained CNN backbone. The main difference between our method and PANDA is how to solve the model collapse issue. In PANDA, it suggests three options: simple early stopping, sample-wise early stopping and continual learning to avoid the model collapse. In our method, we use a stop-gradient strategy to alleviate this issue and we adapt the feature in an dense pixel-wise manner.

To assist the learning of semantic vectors for images, many works~\cite{golan2018deep, bergman2020classification, sohn2020learning, tack2020csi} employed \textbf{self-supervised learning} ~\cite{chen2020simple, he2020momentum, komodakis2018unsupervised, chen2020exploring} to discriminate normal data and outliers. Some methods are predicting rotation of images~\cite{golan2018deep, bergman2020classification} and contrastive learning with usual image augmentation strategies~\cite{sohn2020learning, tack2020csi}. Although these methods well capture the semantic object information in images, they fail to encode the fine-grained local irregularities in anomalies~\cite{li2021cutpaste}. Thus, several works~\cite{devries2017improved, yun2019cutmix, li2021cutpaste} created a set of new data augmentations that replicates the local defects in anomalies. The method Cutout~\cite{devries2017improved} randomly removes a small rectangular area from images, and CutPaste~\cite{li2021cutpaste} further modifies the algorithm to cut a patch from one image and paste it on the other. However, the representation of created negative irregularities usually does not overlap with real-world anomalies~\cite{li2021cutpaste}, which limits the generalization potential of these methods in inference processes. 

In our paper, we follow the method of self-supervision in anomaly detection to propose a novel coarse-to-fine task. Earlier self-supervised methods~\cite{chen2020simple, he2020momentum} typically employ negative samples to learn diversified representations for different classes. Inspired by the recent innovative self-supervised method SimSiam~\cite{chen2020exploring}, we replace the generated anomalous samples that disagree with reality with our non-contrastive learning in a dense pixel-wise self-supervision manner. With the stop-gradient operation~\cite{chen2020exploring} that discards negative samples, we eliminate any prior assumption about the anomalous data in training, and our model can therefore generalize to a variety of anomalies in real world. Furthermore, through our proposed coarse alignment of images and dense supervision of pixel-wise feature learning, we reduce the variances in normal data representation, enabling the learned compact distribution to predict robust distance estimates for outliers.

\begin{figure*}
  \centering
  \includegraphics[width=\linewidth]{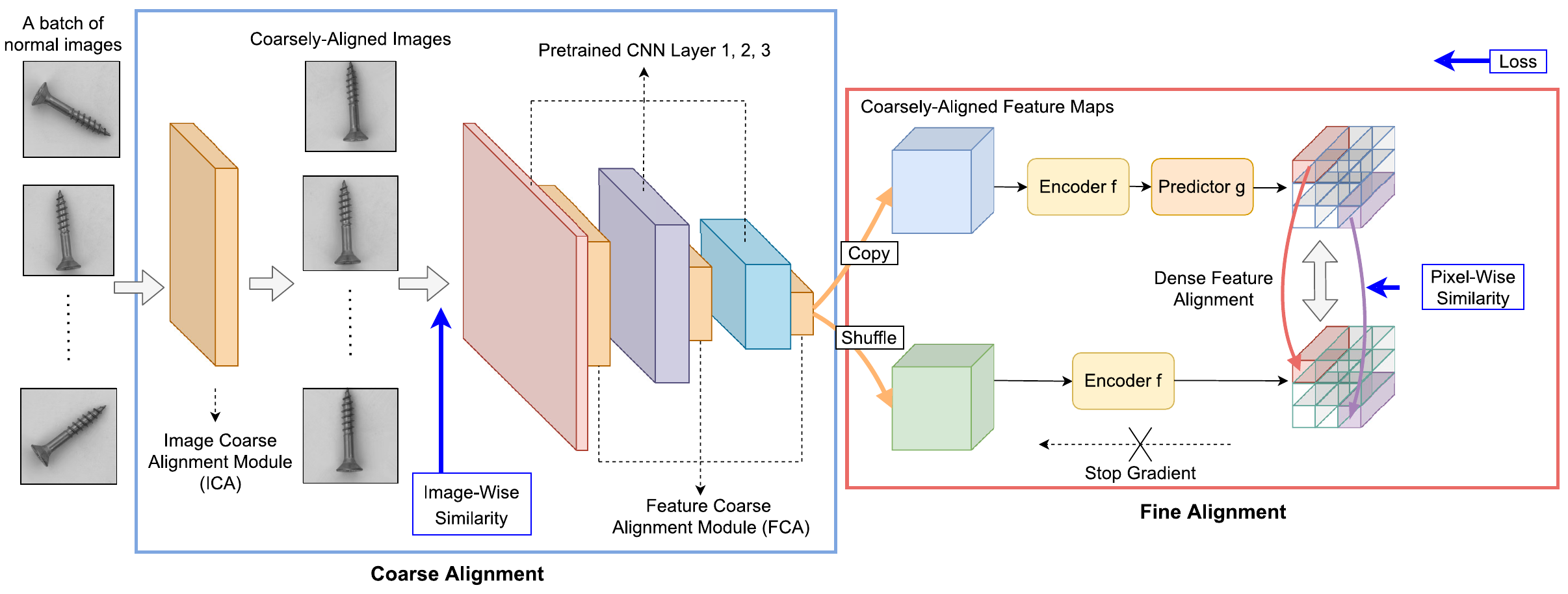}
  \caption{The whole architecture of the proposed approach. The coarse alignment module standardizes the image-wise and feature-wise correspondence among pixels, while the fine alignment module densely maximizes the similarity of these pixels in a batch. We train these two modules with non-contrastive learning method in an end-to-end manner.}
  \label{fig:whole_framework}
\end{figure*}

\section{Coarse-to-Fine Non-Contrastive Learning}
In this section, we demonstrate our novel framework to detect and localize fine-grained anomalies. As indicated in Figure~\ref{fig:whole_framework}, our method consists of a coarse alignment stage and a fine alignment stage. The coarse alignment module first captures and aligns the global context of images. The standardized image features are then passed to the dense representation encoder in for fine-grained self-supervised learning. We construct compact Gaussian distribution from our dense normal features and spot distribution outliers as anomalies.

\subsection{Coarse Alignment Stage}
\label{sec:c2ff}

\subsubsection{Image-Level Coarse Alignment (ICA)}
 The image-level coarse alignment (ICA) aims at regularizing the pixel distribution of normal images: it orients all images in a batch to a similar direction and position for dense comparison. Specifically, we regress the affine transformations $\mathcal{T}_\theta(\mathcal{D}_{i})$ on an input image $\mathcal{D}_{i} \in \mathbb{R} ^ {H \times W \times C}$:
 
 \begin{equation}
   \begin{bmatrix} x_i^t  \\ y_i^t \end{bmatrix} = \mathcal{T}_\theta(\mathcal{D}_i) = \begin{bmatrix} \theta_{11} & \theta_{12} & \theta_{13} \\ \theta_{21} & \theta_{22} & \theta_{23} \end{bmatrix} \begin{bmatrix} x_i^s  \\ y_i^s \\ 1 \end{bmatrix}
   \label{method:affine}
 \end{equation}

 Inspired by Spatial Transformer Network~\cite{jaderberg2015spatial}, we adopt its similar architecture to our ICA module, which uses a tiny network $h_{\mathcal{T}_\theta}$ to learn the above affine mapping from the original image (denoted by $\{(x_i^s, y_i^s)\}$) to a globally aligned representation ($\{(x_i^t, y_i^t)\}$). To train ICA module, we randomly pair the images in a batch and then minimize the $\ell_2$ distance between them:
 
 \begin{equation}
    \mathcal{L}_{ICA} (\mathcal{D}; \theta_h, \mathcal{T}_\theta) = \sum_{A,B \in \mathcal{D}}\sum_{i=0}^{H-1}\sum_{j=0}^{W-1} \lVert h_{\mathcal{T}_\theta}(A_{i,j}) - h_{\mathcal{T}_\theta}(B_{i,j})\rVert_2.
    \label{method:lossica}
 \end{equation} 
 
\begin{figure}[t]
  \centering
  \includegraphics[width=1.\linewidth]{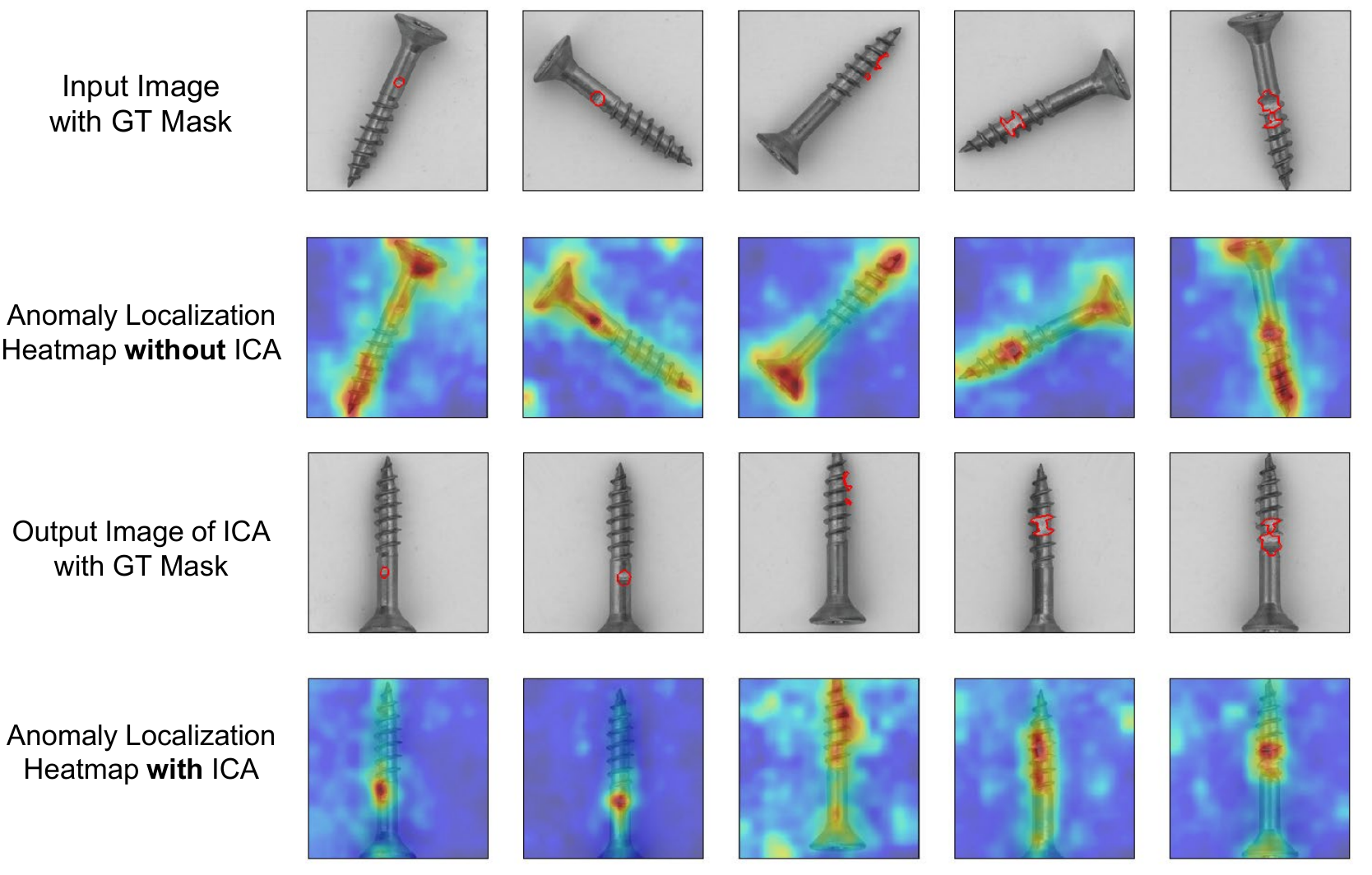}
  \caption{The visualization results for our coarse image alignment module, the top two rows are the results for screws without using ICA and the bottom two rows are the alignment results and anomaly localization results for the same screws with ICA.}
  \label{fig:cam_vis}
\end{figure}

 Note that we do not assign any standard orientation or alignment for our module to regress, so it learns a unified position in a self-supervised manner. Each time we random select two images $A, B \in \mathcal{D}$, the $\ell_2$ loss in Equation~\ref{method:lossica} supervises the alignment of $A$ and $B$ toward the reduction of entropy in this system. Thus, given enough iterations, the system reaches a consensus on alignment and the entropy is thus reduced to a local minimum. The results of roughly unified positions are displayed by the coarsely-aligned images in Figure~\ref{fig:cam_vis}.
 
\subsubsection{Feature-Level Coarse Alignment (FCA)}
  For transformed images in Figure~\ref{fig:cam_vis}, their positions are not strictly aligned to a unified representation. We thus introduce a feature-level coarse alignment (FCA) module to further adjust the high-level image representations. We use a pre-trained ResNet-18~\cite{he2016deep} as the feature extractor. As shown in Figure~\ref{fig:whole_framework}, we insert the FCA between succeeding layers to align the embedding distribution with feature-level affine transformation. FCA has a similar implementation as ICA; nonetheless, this module enforces the backbone to extract generalizable features to align positions in a global high-level embedding. The FCA module is only governed by self-supervised loss in the fine alignment stage. The implementation details of image-level and feature-level alignment are in Appendix.

\begin{algorithm}[t]
  \caption{Pixel-wise Non-contrastive Learning Pseudocode, Pytorch-like}
  \label{alg:pxnl}
    \begin{lstlisting}[language=Python]
# x: features of a minibatch of images (NxCxHxW)
# f: encoder convolution network
# h: predictor convolution network
for x in loader: # load a minibatch x of N samples
    x1 = x.clone() # copy x to get x1
    x2 = x[rand_permutation(N)] # shuffle the batch to generate reordered images as x2
    z1, z2 = f(x1), f(x2) # encodeing
    p1, p2 = g(z1), g(z2) # prediction
    L = D(p1, z2)/2 + D(p2, z1)/2 # loss
    L.backward() # back-propagate
    update(f, g) # SGD update

def D(p, z): # negative cosine similarity
    z = z.detach() # stop gradient
    return - cosine_similarity(p, z, dim=-1).mean()
    \end{lstlisting}
\end{algorithm}

\subsection{Fine Alignment Stage: Pixel-wise Non-contrastive Learning}
\label{sec:fine-alignment-satge}
Non-contrastive learning is the training of a semantically meaningful normal image representation without leveraging its distance with anomalies. To detect the fine-grained anomalies, we propose the pixel-wise alignment module, which maximizes the feature similarity across every embedding position for all normal images.

As indicated by the Algorithm~\ref{alg:pxnl}, we first randomly shuffle the minibatch and sample two feature maps. Let $W, V \in \mathbb{R} ^ {H' \times W' \times C'}$ denote the two encoded image feature map belonging to two different images from the last FCA. Then, for every position $0 \leq i < H'$, $0 \leq j < W'$ in these two features, we extract the corresponding feature vectors $w_{ij} \sim W$ and $v_{ij} \sim V$. We aim at encoding a unique vector representation for each position in the feature map, as well as narrowing its distribution for all normal images. Thus, we use $1\times1$ convolutional operator instead of fully-connected layers in the feature extractor. The vectors are passed to a shared 3-layer \emph{1$\times$1conv} encoder $f$. Only $w_{ij}$ is processed through a 2-layer \emph{1$\times$1conv} predictor $g$ to project its feature to the vector space of $v_{ij}$. Given the two output vectors from the encoder $m_{ij} \triangleq f(w_{ij})$ and $n_{ij} \triangleq f(v_{ij})$, we minimize their negative cosine similarity:

\begin{equation}
    \mathcal{L}_{ij} (m_{ij}, n_{ij}; \theta_g, \theta_f) = - \frac {\langle g(m_{ij}), n_{ij}\rangle } {\lVert m_{ij} \rVert_2 \cdot \lVert n_{ij} \rVert_2},
    \label{method:losscos}
\end{equation} 
where $\theta_g$ and $\theta_f$ are the parameters of encoder and predictor, respectively.
We conduct the above minimization for $w$ and $v$ at all positions $i, j$ respectively to densely supervise the positionally-aligned feature distribution.

To avoid model collapsing~\cite{wang2020understanding} when training with normal data only, we introduce the stop-gradient operation from~\cite{chen2020exploring}. That is, $\nabla \mathcal{L}_{ij}$ is only allowed to descent backward through the upper branch of the network w.r.t. $m_{ij}$, and it updates no information of $n_{ij}$ to the encoder $f$. A symmetry operation is applied to further supervise the learning of robust and generalizable features. The aggregated loss for every position thus becomes:

\begin{equation}
\begin{split}
    \mathcal{L}_{FAS}(m, n; \theta_g, \theta_f) = \sum_{\mathcal{D}}\sum_{i, j} \frac {1} {2} \mathcal{L}_{ij} (m_{ij}, stop\_grad(n_{ij})) +  \\
    \frac {1} {2} \mathcal{L}_{ij} (n_{ij}, stop\_grad(m_{ij})).
\end{split}
\label{method:losssim}
\end{equation}

Its implementation is detailed in Algorithm~\ref{alg:pxnl}. We train the above stages in an end-to-end manner and adjust the weight between coarse and fine alignment with $\lambda_{1}$ and $\lambda_2$. Hence, the final loss in our framework is:

\begin{equation}
    \mathcal{L}_{total}(\, \cdot \,; \theta_{h, f, g}, \mathcal{T}_
    \theta) = \lambda_1  \cdot \mathcal{L}_{ICA} + \lambda_2 \cdot \mathcal{L}_{FAS}.
    \label{method:losstotal}
\end{equation}
$\mathcal{L}_{FAS}$ is the dominant loss function supervising all parameters and $\mathcal{L}_{ICA}$ is the auxiliary loss function used to guarantee the convergence of $ICA$. 
By optimizing the coarse and fine alignment module collectively, we allow the network to self-adjust and learn meaningful correlations of normal image embeddings.

\subsection{Anomaly Score Computation in Inference}
\label{sec:inference}
With the densely extracted features, we model the representation of normal images with the Gaussian distribution for every pixels on feature map following~\cite{defard2020padim}. We extract the normal image representation at position $(i, j)$ by concatenating the three pyramid layers of features of CNN at $(i, j)$. Let $X_{ij}\in \mathbb{R} ^ {(C_1 + C_2 + C_3) \times N}$ denote the aggregated feature from the CNN for all images of training set. We model a distinctive Gaussian distribution $\mathcal{N}(\mu_{ij}, \Sigma_{ij})$ for each pixel $(i, j)$ on the feature map by:
\begin{equation}
    \mu_{ij} = \frac {1} {N} \sum_{k}^{N} x_{ij}^k; \, \Sigma_{ij} = \frac {1} {N - 1} \sum_{k}^{N} (x_{ij}^k - \mu_{ij})(x_{ij}^k - \mu_{ij})^T
    \label{method:normal}
\end{equation}

During inference, we compute the anomaly score by taking the Mahalanobis distances at every pixel between the test images and the normal distribution:
\begin{equation}
    \mathcal{D}(x_{ij}) = \sqrt{(x_{ij} - \mu_{ij})^T\Sigma_{ij}^{-1}(x_{ij} - \mu_{ij})}
    \label{method:score}
\end{equation}
Then, the distance matrix $\mathcal{D}$ is an anomaly map with dense pixel-wise anomaly scores. A greater score indicates a severer anomalous signal. We thus use the maximum anomaly score map to represent the anomaly score for the entire image.

\begin{table*}[tbp]
\begin{center}
\resizebox{\linewidth}{!}{
\begin{tabular}{ccccccccccc}
\toprule
\multicolumn{2}{c}{\multirow{2}{*}{Category}} & DOCC & FCDD & U-S & P-SVDD & SPADE & PaDiM & Cut Paste & Ours & Ours\\
~ & ~ & ~ & ~ & ~ & ~& (WR50) & (WR50) & (R18) & (R18) & (WR50) \\
\toprule
\multirow{6}{*}{texture} & carpet & (90.6, -) & (-, 96) & (95.3, -) & (92.9, 92.6) & (-, 97.5) & (-, 98.9) & (93.1, 98.3) & (98.3$\pm$0.4, 98.4$\pm$0.1) & (\textbf{98.8}$\pm0.2$, \textbf{98.5}$\pm0.1$) \\
~ & grid & (52.4, -) & (-, 91) & (98.7, -) & (94.6, 96.2) & (-, 93.7) & (-, 94.9 & (\textbf{99.9}, \textbf{97.5}) & (97.4$\pm$1.4, 95.7$\pm$0.4) & (98.9$\pm0.8$, 96.8$\pm0.3$)  \\
~ & leather & (78.3, -) & (-, 98) & (93.4, -) & (90.9, 97.4) & (-, 97.6) & (-, 99.1) & (\textbf{100.0}, \textbf{99.5}) & (\textbf{100.0}$\pm$0.0, 98.9$\pm$0.2) & (\textbf{100.0}$\pm0.0$, 99.2$\pm0.1$)  \\
~ & tile & (96.5, -) & (-, 91) & (95.8, -) & (\textbf{97.8}, 91.4) & (-, 87.4) & (-, 91.2) & (93.4, 90.5) & (95.4$\pm$1.1, 93.7$\pm$0.7) & (\textbf{98.8}$\pm0.9$, \textbf{96.8}$\pm0.5$) \\
~ & wood & (91.6, -) & (-, 88) & (95.5, -) & (96.5, 90.8) & (-, 88.5) & (-, 93.6) & (98.6, 95.5) & (\textbf{99.8}$\pm$0.2, 94.2$\pm$0.1) & (99.4$\pm0.4$, \textbf{99.6}$\pm0.2$) \\
\cmidrule(lr){2-11}
~ & average & (81.9, -) & (-, 93) & (95.7, -) & (94.5, 93.7) & (-, 92.9) & (-, 95.6) & (97.0, 96.3) & (98.2$\pm$0.8, 96.2$\pm$0.3) & (\textbf{99.2}$\pm$0.5, \textbf{98.2}$\pm$0.2) \\
\bottomrule
\multirow{11}{*}{object} & bottle & (99.6, -) & (-, 97) & (96.7, -) & (98.6, 98.1) & (-, \textbf{98.4}) & (-, 98.1) & (98.3, 97.6) & (\textbf{100.0}$\pm$0.0, 98.3$\pm$0.0) & (\textbf{100.0}$\pm$0.0, 98.3$\pm$0.0)\\
~ & cable & (90.9, -) & (-, 90) & (82.3, -) & (90.3, 96.8) & (-, 97.2) & (-, 95.8) & (80.6, 90.0) & (94.3$\pm$0.5, 96.7$\pm$0.2) & (\textbf{95.3}$\pm$0.4, \textbf{97.5}$\pm$0.2)  \\
~ & capsule & (91.0, -) & (-, 93) & (92.8, -) & (76.7, 95.8) & (-, \textbf{99.0}) & (-, 98.3) & (\textbf{96.2}, 97.4) & (93.2$\pm$1.8, 98.5$\pm$0.6) & (92.5$\pm$2.0, 98.6$\pm$0.8) \\
~ & hazelnut & (95.0, -) & (-, 95) & (91.4, -) & (92.0, 97.5) & (-, \textbf{99.1}) & (-, 97.7) & (97.3, 97.3) & (99.8$\pm$0.2, 98.3$\pm$0.1) & (\textbf{99.9}$\pm$0.1, 98.7$\pm$0.1) \\
~ & metal nut & (85.2, -) & (-, 94) & (94.0, -) & (94.0, 98.0) & (-, 98.1) & (-, 96.7) & (99.3, 93.1) & (\textbf{99.9}$\pm$0.1, 97.1$\pm$0.2)  & (\textbf{99.9}$\pm$0.1, \textbf{98.2}$\pm$0.1)  \\
~ & pill & (80.4, -) & (-, 81) & (86.7, -) & (86.1, 95.1) & (-, 96.5) & (-, 94.7) & (92.4, 95.7) & (\textbf{94.9}$\pm$1.2, 97.2$\pm$0.3) & (94.5$\pm$1.5, \textbf{97.3}$\pm$0.5) \\
~ & screw & (86.9, -) & (-, 86) & (87.4, -) & (81.3, 95.7) & (-, \textbf{98.9}) & (-, 97.4) & (86.3, 96.7) & (89.7$\pm$0.8, 98.7$\pm$0.5) & (\textbf{90.1}$\pm0.5$, 98.7$\pm$0.4) \\
~ & toothbrush & (96.4, -) & (-, 94) & (98.6, -) & (\textbf{100.0}, 98.1) & (-, 97.9) & (-, 98.7) & (98.3, 98.1) & (99.9$\pm$0.1, \textbf{98.9}$\pm$0.0) & (\textbf{100.0}$\pm$0.0, \textbf{98.9}$\pm$0.0) \\
~ & transistor & (90.8, -) & (-, 88) & (83.6, -) & (91.5, 97.0) & (-, 94.1) & (-, 97.2) & (95.5, 93.0) & (\textbf{99.7}$\pm$0.1, \textbf{98.6}$\pm$0.1)  & (99.2$\pm$0.3, 98.1$\pm$0.3) \\
~ & zipper & (92.4, -) & (-, 95.1) & (95.8, -) & (97.9, 95.1) & (-, 96.5) & (-, 98.2) & (\textbf{99.4}, \textbf{99.3}) & (97.0$\pm$0.5, 97.8$\pm$0.4) & (97.5$\pm$0.5, 98.2$\pm$0.3) \\
\cmidrule(lr){2-11}
~ & average & (90.9, -) & (-, 91) & (90.9, -) & (90.8, 96.7) & (-, 97.6) & (-, 97.3) & (94.3, 95.8) & (96.8$\pm$0.5, 98.0$\pm$0.2)  & (\textbf{97.0}$\pm$0.5, \textbf{98.3}$\pm$0.3) \\
\bottomrule
\multicolumn{2}{c}{average} & (87.9, -) & (-, 92) & (92.5, -) & (92.1, 95.7) & (85.5, 96.5) & (95.3, 96.7) & (95.2, 96.0) & (97.3$\pm$0.5, 97.4$\pm$0.2) & (\textbf{97.7}$\pm$0.4, \textbf{98.2}$\pm$0.3)  \\
\bottomrule
\end{tabular}
}
\end{center}
\caption{Anomaly detection and localization performance on MVTec AD dataset~\cite{bergmann2019mvtec} with the format of (Image-level AUC, Pixel-level AUC). ``WR50" means Wide-ResNet50$\times$2 and ``R18" indicates ResNet-18.}
\label{table:mvtec-results}
\end{table*}

\section{Experiments}
\label{sec:exp}
\subsection{Datasets and Metrics}
We perform experiments on two industrial anomaly detection datastes \textbf{MVTec AD dataset}~\cite{bergmann2019mvtec} and \textbf{BeanTech AD dataset}~\cite{mishra2021vt}.
MVTec AD dataset consists of 5354 real-world images with 15 categories, among which 10 of them are objects and the rest 5 are texture classes. BeanTech AD dataset has 3 categories of 2540 images. In both datasets, the training set consists of only normal images, while the testing set has a mixture of both normal and abnormal images. These datasets provides both anomaly types and anomaly masks as test image labels for evaluation. As mentioned previously, the anomalies in these datasets are more fine-grained than those in the academic dataset settings, e.g. the CIFAR-10~\cite{krizhevsky2009learning} dataset whose anomaly is defined as different object classes. 

Under the one-class classification protocol, we train a model for each category with its respective normal images. The implementation details are listed in Appendix. During inference, we evaluate our method with \textbf{image-level AUC} and \textbf{pixel-level AUC}.

\begin{table}[tpb]
    \centering
    \resizebox{\linewidth}{!}{
    \begin{tabular}{cccccc}
    \hline
         Categories & AE MSE & AE MSE+SSIM & VT-ADL & Ours (Loc) & Ours (Det) \\
         \hline
         01 & 49.0 & 53.0 & 99.0 & 96.1 & 99.6 \\ 
         02 & 92.0 & 96.0 & 94.0 & 95.3 & 95.3\\
         03 & 95.0 & 89.0 & 77.0 & 99.7 & 99.5\\
         \hline
         average & 78.0 & 79.0 & 90.0 & 97.0 & 98.1 \\
    \hline
    \end{tabular}}
    \caption{Anomaly detection and localization performance on BeanTech AD dataset.}
    \label{tab:beantech-results}
\end{table}
\begin{table*}[t]
    \centering
    \resizebox{\linewidth}{!}{
    \begin{tabular}{c|cccc|cccc}
        \toprule
         ~ & \multicolumn{4}{c}{Color Transformation} & \multicolumn{4}{c}{Spatial Transformation} \\
         \cmidrule(lr){2-9}
         ~ & Baseline (det) & Ours (det) & Baseline (loc) & Ours (loc) & Baseline (det) & Ours (det) & Baseline (loc) & Ours (loc) \\
        \hline
         carpet & 98.5 & 99.3 & 97.4 & 98.7 & 98.1 & 97.3 & 97.9 & 97.7 \\
         grid & 90.1 & 95.5 & 93.1 & 94.6 & 91.8 & 95.4 & 93.2 & 95.2 \\
         leather & 99.5 & 100.0 & 98.8 & 98.9 & 99.8 & 100.0 & 97.8 & 98.6 \\
         tile & - & - & - & - & 96.2 & 97.3 & 91.5 & 94.4 \\
         wood & - & - & - & - & 98.7 & 99.1 & 92.9 & 93.8 \\
         bottle & 98.9 & 99.9 & 97.2 & 98.1 & 99.6 & 99.7 & 97.4 & 97.5 \\
         cable & 92.8 & 95.7 & 96.6 & 97.2 & 93.4 & 96.6 & 95.7 & 96.5\\
         capsule & - & - & - & - & 86.8 & 87.9 & 97.5 & 97.6 \\
         hazelnut & 94.2 & 96.4 & 97.1 & 97.4 & 98.2 & 99.8 & 97.5 & 97.7\\
         metal nut & 99.1 & 99.1 & 96.1 & 96.2 & 96.8 & 97.7 & 94.5 & 95.2\\
         pill & 90.8 & 92.2 & 94.3 & 95.8 & 89.7 & 92.2 & 94.8 & 96.9 \\
         screw & 89.7 & 90.9 & 98.1 & 98.3 & 66.8 & 70.9 & 95.2 & 97.4 \\
         toothbrush & 99.6 & 99.9 & 98.5 & 98.4 & 99.8 & 98.7 & 98.1 & 97.9 \\
         transistor & 98.7 & 98.9 & 97.7 & 98.1 & 92.8 & 94.0 & 92.3 & 96.9 \\
         zipper & 92.5 & 96.9 & 97.5 & 97.9 & 90.3 & 97.8 & 97.8 & 97.3 \\
         \hline
         average & 95.4 & 97.1 & 96.7 & 97.5 & 93.3 & 95.6 & 95.5 & 96.7 \\
         \bottomrule
    \end{tabular}}
    \caption{Anomaly detection and localization performance on Disturbed MVTec AD dataset with ResNet-18, ``det" is image-level AUC and ``loc" is pixel-level AUC.}
    \label{tab:disturbed-mvtec}
\end{table*}
\subsection{Anomaly Detection and Localization for MVTec}
 In Table~\ref{table:mvtec-results}, we compare our method with the state-of-the-art one-class anomaly detection approaches in MVTec AD dataset, including deep one-class classifier (DOCC)~\cite{ruff2021unifying},  FCDD~\cite{liznerski2020explainable}, uninformed student (U-S)~\cite{bergmann2020uninformed}, patch SVDD~\cite{yi2020patch}, SPADE~\cite{cohen2020sub}, PaDiM~\cite{defard2020padim}, Cut Paste~\cite{li2021cutpaste} under the metrics of image-level AUC and pixel-level AUC, we give the results of the mean and standard deviation of 5 repeated experiments. With our proposed coarse-to-fine non-contrastive learning method, we achieve the best result among all existing works and make notable improvements on both of texture and object defects. Our method surpasses the current state-of-the-art by a margin of 2.1, yieding 97.7 image-level AUC and 98.2 pixel-level AUC. Some results of anomaly localization are visualized in Fig~\ref{fig:result_vis} for our method, and more comprehensive results of defect localization are provided in Appendix. We can observe that not only does the anomaly heatmap highlights the object with defect, it also displays intense and fine-grained attention on the small anomalous regions. This proves how our framework focuses precisely on the anomalies.
\subsection{Anomaly Detection and Localization for BeanTech}
 In Table~\ref{tab:beantech-results}, we compare our method with the anomaly detection approaches reported in ~\cite{mishra2021vt} on the BeanTech AD dataset. They applied auto-encoder with MSE loss, auto-encoder with MSE and SSIM loss, and VT-ADL in anomaly localization. We give the results of ResNet-18 and we additionally report the image-level results. Our method achieves the best result among all existing and surpasses the current state-of-the-art by a margin of 7.0, yielding 98.1 image-level AUC and 97.0 pixel-level AUC. This result shows our method's potential to generalize to new anomalous detection scenarios, where the anomalous data has varied distribution and needs close scrutinization of details.

\subsection{Performance for Disturbed MVTec}
Considering that the current MVTec AD dataset does not contain products with multiple appearances but only has spatially aligned products, we build a new dataset called Disturbed MVTec through various augmentations. We aim at simulating more challenging real life detection situations and verify the robustness and effectiveness of our framework. Specifically, we build two disturbed scenarios with color and spatial transformation, respectively. For the color transformation, we apply random brightness contrast enhancement and limited adaptive histogram equalization in MVTec's train and test images. For the spatial transformation, we apply random zoom in, zoom out, rotation and flip in the train and test images in MVTec. The examples of these two disturbed datasets and augmentation details are documented in our Appendix. Corresponding experimental results wth ResNet-18 are reported in Table~\ref{tab:disturbed-mvtec}. It should be noted that some categories are suitable for color augmentation, especially for the texture classes. For instance, areas of different colors will be treated as anomalies on tile, wood and capsule, so we did not perform color augmentation on them. Moreover, some categories may not suitable do some special kinds of spatial augmentation. For instance, we can not apply vertical flip or rotation in transistor. We can observe that our method performs generally better than the baseline~\cite{defard2020padim} under these color and spatial transformations, which shows our framework's adaptability to complicated settings that simulate the real-world disturbances in industrial images.

\subsection{Quantitative Analysis on Distribution}
To quantify the improvement of distribution compactness in our work, we compare the distance variance score of our method to the baseline (ImageNet pre-trained model). Given a class of total N normal image representations $X \in \mathbb{R} ^ {N \times H \times W \times C}$, the distance variance score is calculated by first computing each pixel's distance to its alignment center:
\begin{equation}
    \mu_{ij} = \frac {\sum_n^N x_{nij}} {N}, \forall \, 0 \leq i < H, 0 \leq j < W
\end{equation}
\begin{equation}
    D_{nij} = \sqrt{(x_{nij} - \mu_{ij})^T\Sigma_{ij}^{-1}(x_{nij} - \mu_{ij})}
\end{equation}
Then, we get the distance map $D \in \mathbb{R} ^ {N \times H \times W }$ at every location for all normal images. We compute the distance variance score as the aggregated variance for all distance embedding:
\begin{equation}
    \mathcal{S}_{cls} = \frac {1} {H \times W} \sum_{i,j} \sum_n^N \frac {(D_{nij} - \overline{D}_{ij})^2} {N},
\end{equation}

where $\overline{D}_{ij}$ is the average distance for all images at position $(i, j)$.
If the distribution of image representations becomes more compact, then $\mathcal{S}_{class}$ would decrease, since the variation in distances between feature vectors to their center should be smaller. We finally aggregate the distance score for all 15 categories in our dataset:
\begin{equation}
    \mathcal{S}_{total} = \frac {1} {N_{cls}} \sum_{cls} \mathcal{S}_{cls}
\end{equation}

The $\mathcal{S}_{total}$ for the baseline is 0.95, and our method achieves a score of \textbf{0.84}, which is a 11.6\% decrease in the variance score. It thus demonstrates that our method effectively shrinks the feature distributions, while it also keeps the meaningful and discriminating variations in encoding.
\begin{figure*}[tbp]
  \centering
  \includegraphics[width=\textwidth]{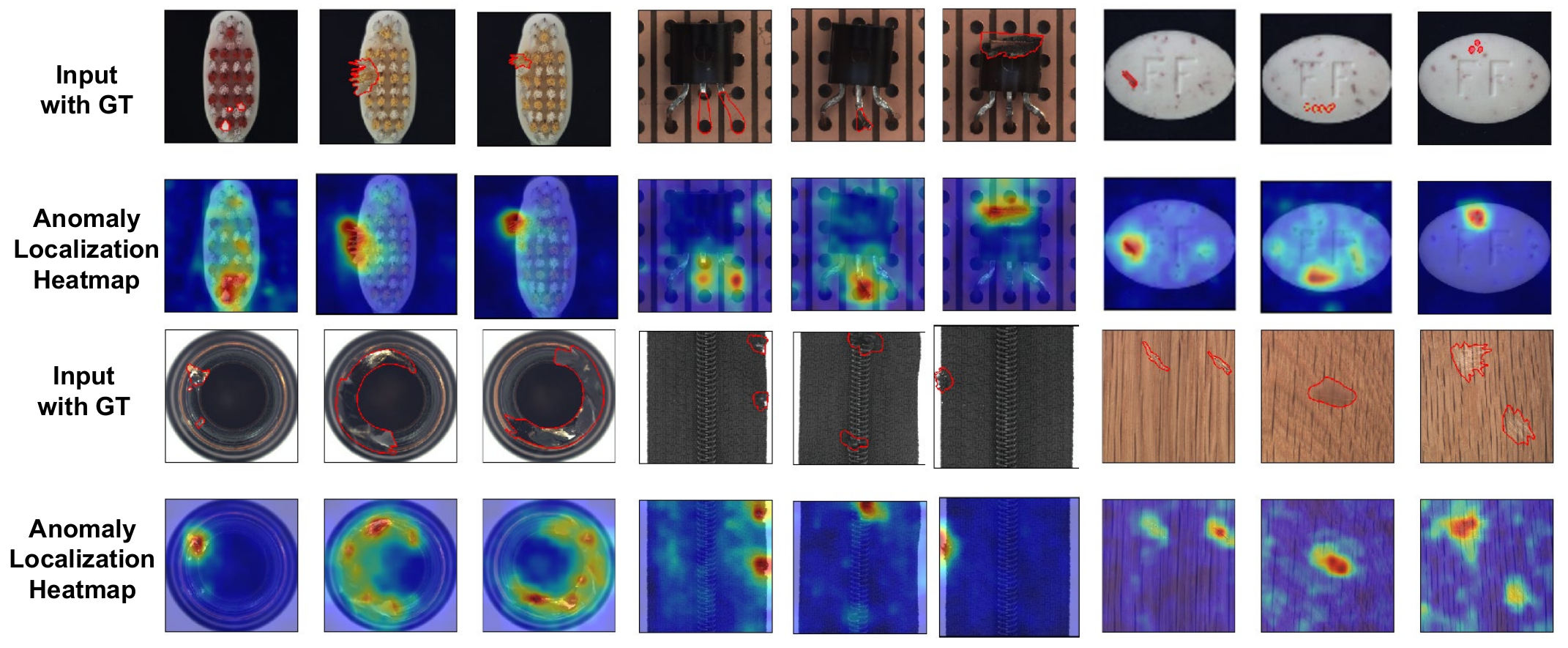}
  \caption{Defect localization on toothbrush, transistor, pill, bottle, zipper, and wood classes of MVTec AD datasets. From top to bottom, input images with ground-truth localization area labeled in red and anomaly localization heatmaps.}
  \label{fig:result_vis}
\end{figure*}
\section{Ablation Study}
\subsection{Component-wise Analysis}
\label{sec:component-wise-analysis}
We investigate the contributions of the main components for our method in Table~\ref{table:component-wise-results}. ``Baseline" only uses the ImageNet pre-trained ResNet-18 to model the Gaussian distribution in Equation~\ref{method:normal} in inference. ``Non-contrastive Learning" is the fine alignment stage to densely maximize the similarity of all images in a batch with the stop-gradient operation. ``Coarse Alignment" is the initial image-level and feature-level alignment modules for distribution regularization. The baseline gives less AUC and PRO results comparing to other methods, such as U-S~\cite{bergmann2020uninformed} and PaDiM~\cite{defard2020padim}. Adding a single non-contrastive learning block improves the image-level AUC to 95.7 and pixel-level AUC to 96.8, which surpasses all previous works. This demonstrates the effectiveness of our designed non-contrastive learning module, because it eliminates the abnormal samples in training and shrinks the distribution of normal samples. Then, adding the coarse alignment module further enhances our advantage over the current state-of-the-art. This is very intuitive, since without first aligning the coarse locations of images, the densely minimized distance among pixels may not be correctly associated. Hence, the ablation study shows the additive effect of each module and the comprehensiveness of our framework.

\begin{table}[tbp]
\begin{center}
\resizebox{\linewidth}{!}{
\begin{tabular}{ccccccc}
\toprule
\multirow{2}{*}{Baseline} & \multirow{2}{*}{Dense Non-contrastive Learning} & \multicolumn{2}{c}{Coarse Alignment} & \multirow{2}{*}{Image-level AUC} & \multirow{2}{*}{Pixel-level AUC} & \multirow{2}{*}{PRO} \\
~ & ~ & ICA & FCA & & & \\
\cmidrule(lr){1-4} \cmidrule(lr){5-7}
\checkmark & & & & 92.3 & 95.4 & 89.1 \\
\checkmark & \checkmark & &  & 95.7 & 96.8 & 90.5  \\
\checkmark & \checkmark & \checkmark & & 96.2 & 97.0 & 90.7  \\
\checkmark & \checkmark & \checkmark & \checkmark & 97.3 & 97.4 & 91.8  \\
\bottomrule
\end{tabular}
}
\caption{Ablation study for effects of each component.}
\label{table:component-wise-results}
\end{center}
\end{table}

\begin{table}[tbp]
\begin{center}
\resizebox{\linewidth}{!}{
\begin{tabular}{c|cccccccc}
\toprule
~ & carpet & grid & leather & tile & wood & bottle & cable & capsule \\
w/o & \textbf{99.3} & 96.3 & 99.5 & \textbf{95.9} & 98.8 & \textbf{100.0} & 89.4 & 90.4 \\
w & 98.3 & \textbf{97.4} & \textbf{100.9} & 95.4 & \textbf{99.8} & \textbf{100.0} & \textbf{94.3} & \textbf{93.2} \\
\hline
~ & hazelnut & pill & metal nut & screw & toothbrush & transistor & zipper & average \\
w/o & 96.1 & 99.1 & 91.8 & 74.2 & 99.4 & 94.8 & 89.5 & 94.3 \\
w & \textbf{99.8} & \textbf{99.9} & \textbf{94.9} & \textbf{89.7} & \textbf{99.9} & \textbf{99.7} & \textbf{97.0} & \textbf{97.3}\\
\bottomrule
\end{tabular}
}
\caption{The class-wise image-level AUC for our method with or without coarse alignment module.}
\label{table:coarse_class_wise}
\end{center}
\end{table}
\subsection{Effects of Coarse Alignment Stage}
We give the qualitative results of a specific class to show the effects of the coarse alignment stage in Table~\ref{table:coarse_class_wise}. We can observe that the coarse align stage can improve most of the categories in MVTec AD dataset, especially for ``screw", and we will discuss this improvement is achieved below.
\subsubsection{Image-Level Coarse Alignment module}
To visualize the result of image-level coarse alignment, we choose the most disordered category screw. As indicated in Figure~\ref{fig:cam_vis}, the screws in dataset have different orientations and positions. The localization heatmap has sparse and distributed attention over the screws, which cannot accurately localize the defects. After passing them through ICA, they are all roughly aligned to the straight-up direction. The heatmap also becomes more focused on the specific defect locations, instead of spreading in different orientations and positions. That is to say, the ICA module narrows the distribution of normal samples and reduces the difficulties in feature learning, which improves the anomaly detection and localization performance. 

\begin{table}[t]
\begin{center}
\resizebox{\linewidth}{!}{
\begin{tabular}{cccccccc}
\toprule
Baseline & Layer1 & Layer2 & Layer3 & Image-level AUC & Pixel-level AUC & PRO \\
\cmidrule(lr){1-4} \cmidrule(lr){5-7}
\checkmark & & & & 96.2 & 96.4 & 90.4 \\
\checkmark & \checkmark &  & & 96.6 & 96.5 & 90.5 \\
\checkmark & & \checkmark &  & 96.7 & 96.6 & 90.7  \\
\checkmark & & ~ & \checkmark & 96.5 & 96.2 & 90.5  \\
\checkmark & \checkmark & \checkmark & \checkmark & 97.3 & 97.4 & 91.8  \\
\bottomrule
\end{tabular}
}
\caption{Effects of the insert location of feature-level coarse alignment module.}
\label{table:cfa}
\end{center}
\end{table}
\subsubsection{Feature-Level Coarse Alignment Module}
We investigate the effects of the position of the feature-level coarse alignment module in Table~\ref{table:cfa}. The baseline is the clean backbone without feature-level alignment. We find that inserting FCA after a single feature layer has limited improvement over the baseline, while inserting it in all three layers gives a thorough boost in AUC and PRO. We speculate that although adding FCA in a single layer enables the network to adjust the feature's positions, it limits the flexibility and reception field of the alignment to a single scope. Adding them to all three layers successively reinforces the feature alignment process and allows the self-supervised signal in Equation~\ref{method:losssim} to descent backward without information loss.

\section{Conclusion}
In this paper, we propose a coarse-to-fine non-contrastive learning framework for unsupervised anomaly detection. The key to our success is the dense non-contrastive learning with coarse alignment and fine alignment module, which encourages the model to learn and narrow down the distribution of normal patterns. Our method achieves high performance on the industrial defect dataset and surpasses the state-of-the-art in both anomaly detection and localization tasks. 

\clearpage
\twocolumn[
\begin{@twocolumnfalse}
	\section*{\centering{Supplementary Material for \\ \emph{Focus Your Distribution: Coarse-to-Fine Non-Contrastive Learning for AnomalyDetection and Localization\\[25pt]}}}
\end{@twocolumnfalse}
]

\begin{figure*}[t]
  \centering
  \includegraphics[width=\textwidth]{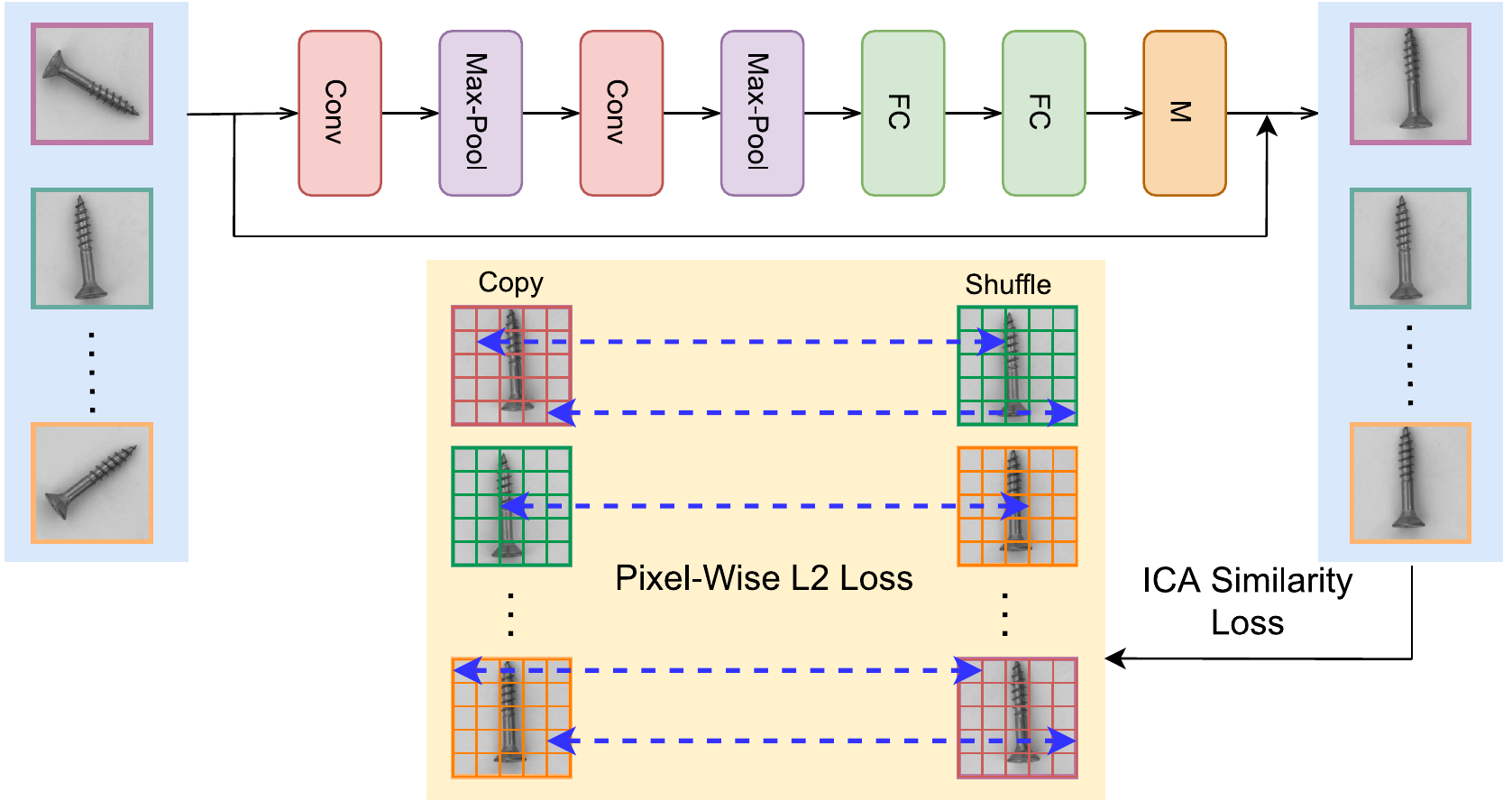}
  \caption{The architecture of Image-level Coarse Alignment module. ICA, supervised by similarity loss, learns the rotation transformation parameters from input images and then applies the affine matrix to them to get batch images in same pose.}
  \label{fig:ica}
\end{figure*}

\section{Details on Experiments}
\subsection{Implementation Details of two Coarse Alignment modules}
Our implementation of the Image-level Coarse Alignment module (ICA) and Feature-level Coarse Alignment module (FCA) is inspired by~\cite{jaderberg2015spatial}. As shown in Figure~\ref{fig:ica}, ICA contains two $3\times3$ Convolution layers, two max-pooling layers, and two fully-connected layers. The final FC layer outputs the angle for rotation transformation matrix of input images, which then aligns them to a unified direction.

We develop a self-supervised learning task to train the ICA. Specifically, the ICA loss minimizes the $\ell_2$ distance between each paired images on one batch in a pixel-wise manner. The pixels are thus enforced to learn shared, representitive position alignment information. We give the details of this self-supervised learning task and its loss function in Algorithm~\ref{alg:pwl2}.

FCA shares the same the basic structure with ICA, and the only difference between them is the transformation matrix M. In ICA, the matrix M is only used to rotate the image with a single rotation angle, while the M in FCA contains 6 parameters for the affine transformation, including scale, rotation, and translation. Moreover, different from ICA who has a distinctive $\ell_2$ similarity loss, the supervision signal for FCA is from the non-contrastive learning loss in the fine alignment stage. Therefore, the feature-wise fine alignment gives a full set of high-level alignment, which facilitates the adjustment of features in its downstream non-contrastive learning.


\begin{algorithm}[htb]  
  \caption{Pixel-wise Image Coarse Alignment Pseudocode, Pytorch-like}
  \label{alg:pwl2}
    \begin{lstlisting}[language=Python]
    # x: features of a minibatch of images (NxCxHxW)
    # I: ICA network
    for x in loader: # load a minibatch x of N samples
        x1 = x.clone() # copy x to get x1
        x2 = x[rand_permutation(N)] # shuffle the batch to generate reordered images as x2
        z1, z2 = I(x1), I(x2) # coarse align
        L = l2(z1, z2)# pixel-wise l2 loss
        L.backward() # back-propagate
        update(I) # SGD update
    \end{lstlisting}
\end{algorithm} 

\subsection{Implementation Details of Training}
We use the first three blocks of ImageNet pre-trained ResNet-18~\cite{he2016deep} as our backbone network. We train our model on $224\times224$ image with one GPU. We update the parameters using momentum SGD with the learning rate of 0.01 for the ICA and 0.0001 for the others. The momentum is set to 0.9, and the batch size is 32. Moreover, we use a single cycle of cosine learning rate decay schedule and L2 weight regularization with a coefficient of 0.00001.

\subsection{Implementation Details of Disturbed MVTec}
\begin{figure}[thp]
  \centering
  \includegraphics[width=0.8\linewidth]{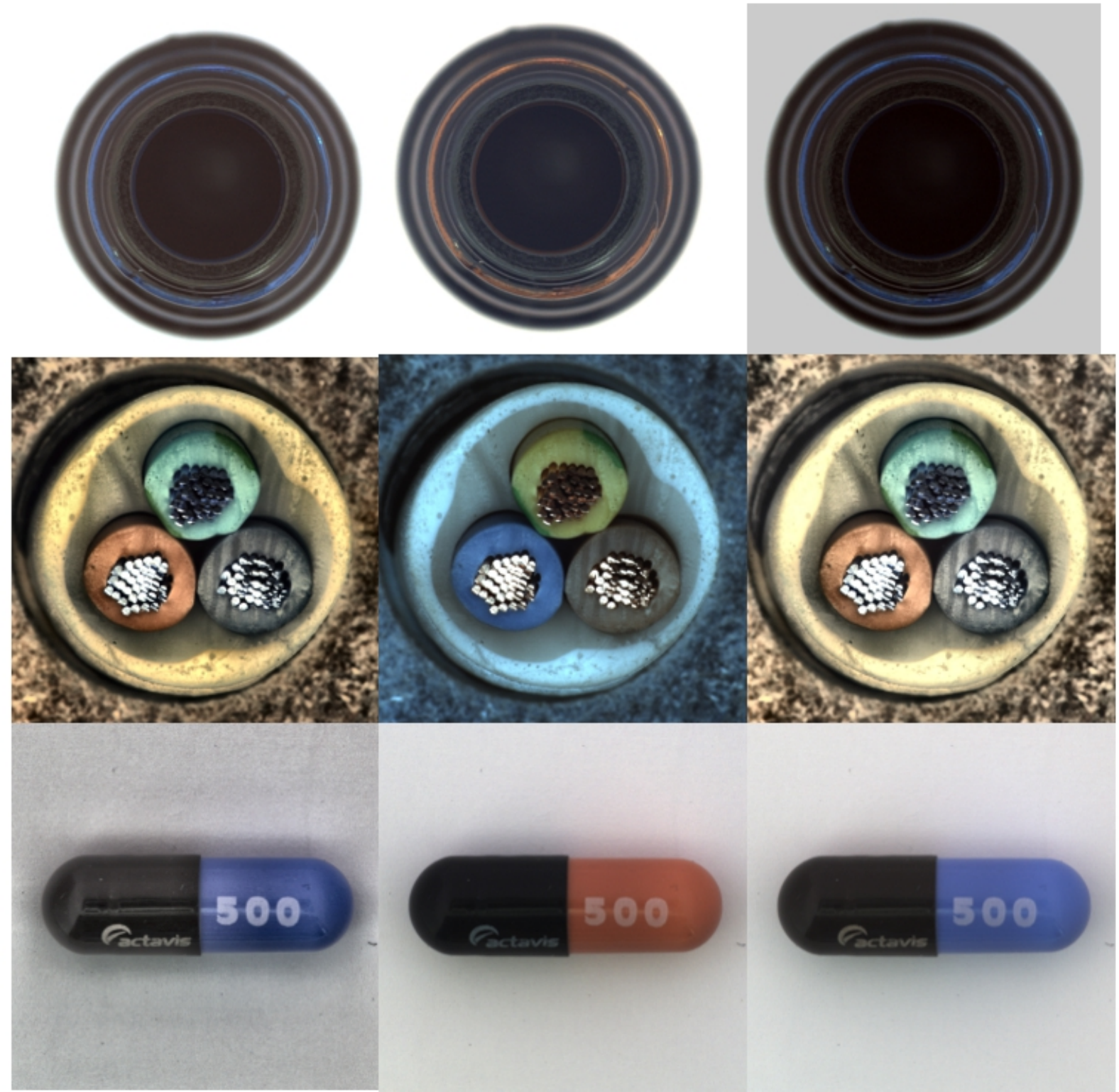}
  \caption{Some image examples of color transformation in MVTec AD dataset.}
  \label{fig:color}
\end{figure}

For the color transformation, we apply random brightness contrast enhancement and limited adaptive histogram equalization in the train and test images in MVTec. See the examples in Figure~\ref{fig:color}. We train our model without ICA and FCA in this new color-transformed dataset, the results are reported in the below table. It should be noted that some categories may not suitable do color augmentation, especially for the texture classes. For instance, areas of different colors will be treated as anomalies on tile, wood and capsule, so we did not experiment on them.

\begin{figure*}[thp]
  \centering
  \includegraphics[width=0.9\textwidth]{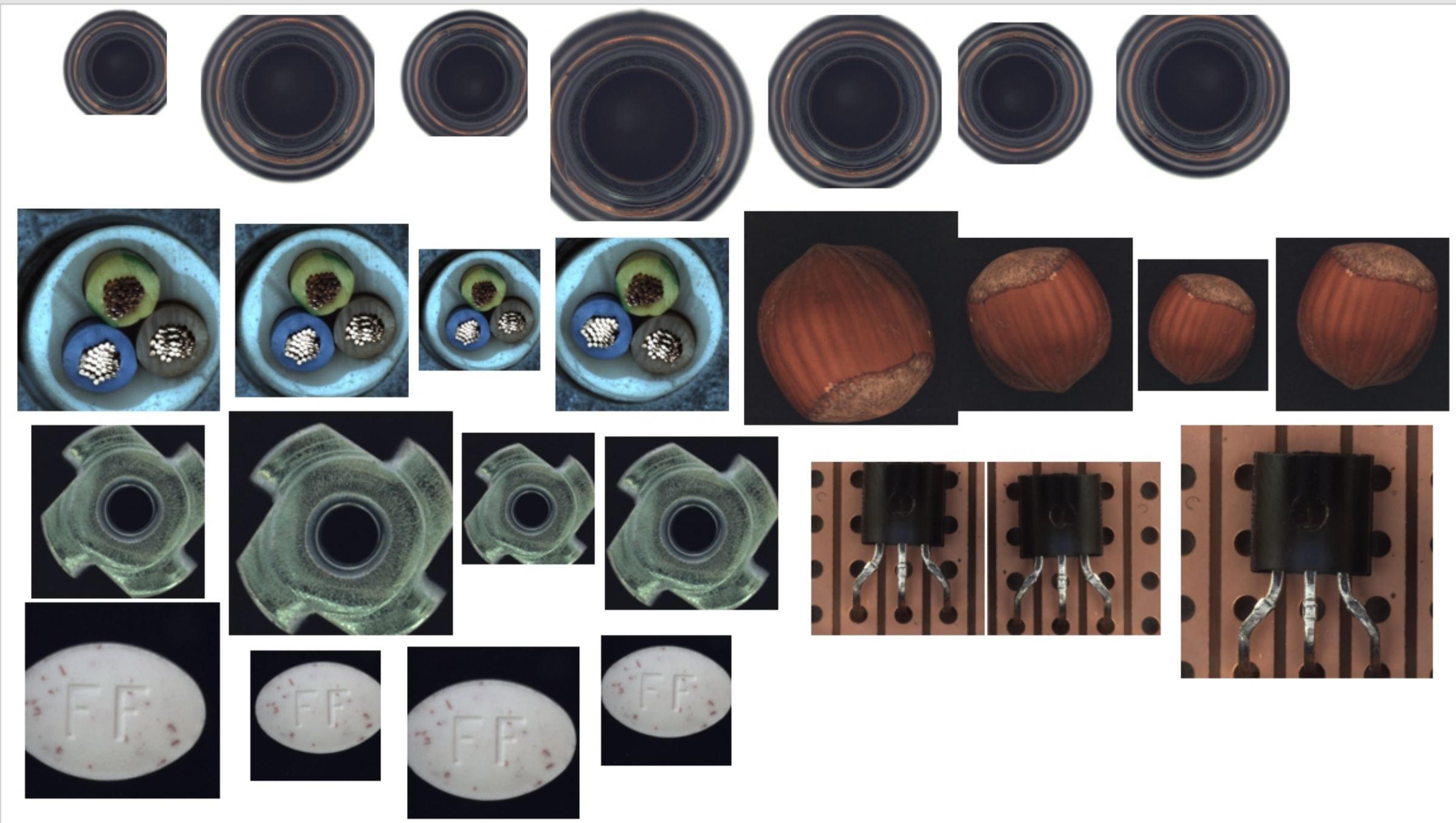}
  \caption{Some image examples of spatial transformation in MVTec AD dataset.}
  \label{fig:spatial}
\end{figure*}

For the spatial transformation, we apply random zoom in/out/rotation/flip in the train and test images in MVTec. See the examples in Figure~\ref{fig:spatial}. We train our model without ICA and FCA in this new spatial-transformed dataset, the results are reported in the below table. It should be noted that some categories may not suitable do some special kinds of augmentation. For instance, we can not apply flip or rotation in transistor, because transistors are required to be arranged in the specified direction and position.

\section{Optimization Hypothesis in Fine Alignment Stage}

We introduce the stop-gradient operation from~\cite{chen2020exploring} to allow training with only normal data and avoid collapsing. We now provide a hypothesis for the mechanism in this non-contrastive learning framework, especially how it helps our model to fit and align compact distribution of normal images.

\subsection{Optimization Problem Formulation}

We hypothesize that our non-contrastive learning implicitly defines a nonparametric clustering algorithm, and the stop-gradient operation thus becomes a decent procedure in the optimization steps. The purpose of our non-contrastive learning module is to regress a representative embedding for a class of normal images along the pixel dimension. This can then be viewed as a single-mode seeking problem for all normal data, from which we identify the most typical features as the mode. At the best case, the mode is the ground truth representation $\mu_A$ for all normal objects in class $A$. 

A natural algorithm for mode seeking is the mean-shift method proposed by~\cite{comaniciu2002mean}. That is, let $x_i \in \mathbb{R}^D, i = 1, 2, ...N$ denote N independent random variables, the mean-shift updates $x$ through a fixed-point iteration:
 \begin{equation}
 x_i = \mathbf{m}(x_i) = \frac {\sum_j^N x_j K(x_i, x_j)} {\sum_j^N K(x_i, x_j)},
 \label{meanshift}
 \end{equation}
where $K(x_i, x_j)$ is the kernel that weighs the distance between $x_i$ and $x_j$. The update continues as $\mathbf{m}(x)$ converges.

In our setting, let $m_{ij} \sim \mathcal{M} \in \mathbb{R}^{H \times W \times C}$ and $n_{ij} \sim \mathcal{N} \in \mathbb{R}^{H \times W \times C}$ denote the two corresponding pixels sampled from the feature map $\mathcal{M}$ and $\mathcal{N}$. For each iteration $t$, we define the loss as: 
\begin{equation}
    \mathcal{L} = \min_{\theta} \mathcal{D} (m_{ij}, n_{ij})  = \mathcal{D}( g^{(t)}(f^{(t)}(m_{ij})), f^{(t)}(n_{ij})),
\end{equation}

where $\theta$ are parameters for the predictor $g$ and encoder $f$.

We use negative cosine similarity as $\mathcal{D}(m_{ij}, n_{ij})$ in our paper, but it can be relaxed to any distance measurements. To regress the global minimizer where $g^*(f^*(m_{ij})) = f^*(n_{ij})$ for all $m_{ij}, n_{ij}$, we perform the gradient descent update of $\mathcal{L}$:

\begin{equation}
    \theta^{(t+1)} = \theta^{(t)} + \eta \frac{\partial \mathcal{L}}{\partial m_{ij}}
\end{equation}

With the stop-gradient operation, $n_{ij}$ is treated as a constant and does not back-propagate its gradient to $\mathcal{L}$. We can thus interpret $f^{(t)}(n_{ij})$ as the optimizing target for $f$ and $g$ to regress toward. Therefore, at each iteration $t$, we update new parameters by treating previous $f^{(t)}(n_{ij})$ as the approximation for the ground truth representation $\mu_A$:

\begin{equation}
    \theta_g^{(t+1)}, \theta_f^{(t+1)} \leftarrow \min \mathcal{D}(\, \cdot \,, f^{(t)}(n_{ij}))
\end{equation}

As we aggregate the updates for a batch of images from our random shuffle algorithm, the process of aligning each pair of $m_{ij}$ and $n_{ij}$ toward $f^{(t)}(\, \cdot \,)$ is equivalent to:
\begin{equation}
    g^{(t+1)}(f^{(t+1)}(\, \cdot \,)) \simeq \mathbb{E}_Z[f^{(t)}(Z)],
    \label{expect}
\end{equation}

where Z is the random variable for normal image feature maps in a batch. Since we perform the pairing of images by random and assign equal weights to all images, Eq~(\ref{expect}) becomes:
\begin{equation}
    g^{(t+1)}(f^{(t+1)}(\, \cdot \,)) \simeq \mathbb{E}_Z[f^{(t)}(Z)] = \frac {\sum_i f^{(t)}(z_i)} {|B|},
    \label{batch_exp}
\end{equation}
where $|B|$ denotes the batch size. We can relax Eq~(\ref{batch_exp}) to Eq~(\ref{meanshift}) by adding a flat kernel:

\begin{equation}
 \mathcal{K}(\, \cdot \, , z_i) =
\begin{cases}
1, &  z_i \in B \\
0, &  \textup{otherwise}
\end{cases}, 
\end{equation}

where $B$ represents all feature maps in a batch. Then, 

\begin{equation}
    g^{(t+1)}(f^{(t+1)}(\, \cdot \,)) \simeq \mathbb{E}_Z[f^{(t)}(Z)] = \frac {\sum_i f^{(t)}(z_i) \mathcal{K}(\, \cdot \, , z_i)} {\sum_i \mathcal{K}(\, \cdot \, , z_i)}
\end{equation}

Our non-contrastive learning thus becomes an implicitly-defined mean shift algorithm. At each iteration, among all normal images in the training set, the parameters of embedding only shift toward the mean of randomly sampled feature maps from a single batch. By the convergence of fixed-point iteration~\cite{burden2011numerical}, our method regresses to a mode representation of all normal images given enough training epoches. The scattered feature vectors for embedding thus shrink into a compact distribution where their feature encodings are shared and aligned.

\section{Experimental Results of PRO Metric}
The pixel-level AUC score favors over large anomalies. To resolve this, Bergmann et al~\cite{bergmann2019mvtec} introduced the PRO (per-region overlap) metric. For each connected component in the anomaly mask, it plots the mean correctly classified pixel rates over the false positive rate (FPR). The PRO score is the normalized value of the integral of this curve from 0 to 0.3 FPR. A greater PRO score indicates better performance in localizing both obvious and subtle anomalies. We compare our method with existing works under per-region-level in Table~\ref{table:pro}. We can observe that our method surpasses the state-of-the-art.

\section{Application to Semantic Outlier Detection}
We conduct the semantic anomaly detection experiment
on CIFAR-10~\cite{krizhevsky2009learning} following the protocol in~\cite{golan2018deep,sohn2020learning},
where a single class is treated as normal and the remaining 9
classes are anomalies. We achieves 66.7 AUC, which surpasses some previous works, including OCSVM~\cite{scholkopf2001estimating}, KDE~\cite{parzen1962estimation}, AnoGAN~\cite{schlegl2017unsupervised}, DeepSVDD~\cite{ruff2018deep} and OCGAN~\cite{perera2019ocgan}. Still, its performance lags behind some other algorithms that specifically designed for image-level anomaly detection~\cite{perera2019ocgan}. This result highlights the difference between fine-grained anomaly detection and image-level defect detection. While the image-level detection focuses on the overall semantic information, our fine-grained anomaly detection aims at detecting subtle anomalies that are usually indiscernible in global semantic context. Such difference suggests that these two detection tasks need different algorithms to target different aspects for the problem.

\section{Anomaly Detection and Localization for Shanghai Tech Campus Dataset}
To further investigate the generalization performance of our method, we conduct experiments on Shanghai Tech Campus Dataset. Considering Shanghai Tech is used for video anomaly detection which has 13 sences, we choice 8 sences from them which don't require the correlation information of the front and rear frames of the video to detect anomaly. We draw images every 5 frames and use all the extracted single frame for training. The results are shown in Table~\ref{tab:shanghai}.


\begin{table}[tbp]
\begin{center}
\resizebox{\linewidth}{!}{
\begin{tabular}{cccc}
\toprule
\multirow{2}{*}{Category} & U-S~\cite{bergmann2020uninformed} & PaDiM~\cite{defard2020padim} & Ours\\
 ~ & ~ & (WR50) & (WR50)  \\
\toprule
texture & 79.4 & \textbf{93.2} & 93.1  \\
object & 88.9 & 91.3 & \textbf{93.0}\\
average &  85.7 & 92.1 & \textbf{93.0}  \\
\bottomrule
\end{tabular}
}
\caption{Anomaly localization performance (PRO) on MVTec AD dataset~\cite{bergmann2019mvtec}. ``WR50" means Wide-ResNet50$\times$2.}
\label{table:pro}
\end{center}
\end{table}

\begin{table}[t]
    \centering
    \resizebox{\linewidth}{!}{
    \begin{tabular}{ccccccccc}
    \toprule
         Sence & 01 & 02 & 03 & 06 & 09 & 10 & 11 & average\\
         \hline
         Pixel-level AUC & 98.8 & 99.0 & 97.5 & 99.4 & 96.0 & 98.1 & 93.9 & 97.5\\
    \bottomrule
    \end{tabular}}
    \caption{Anomaly localization performance (Pixel-level AUC) on Shanghai Tech Campus Dataset.}
    \label{tab:shanghai}
\end{table}

\section{More Anomaly Localization Visualizations}
\begin{figure*}[thp]
  \centering
  \includegraphics[width=\textwidth]{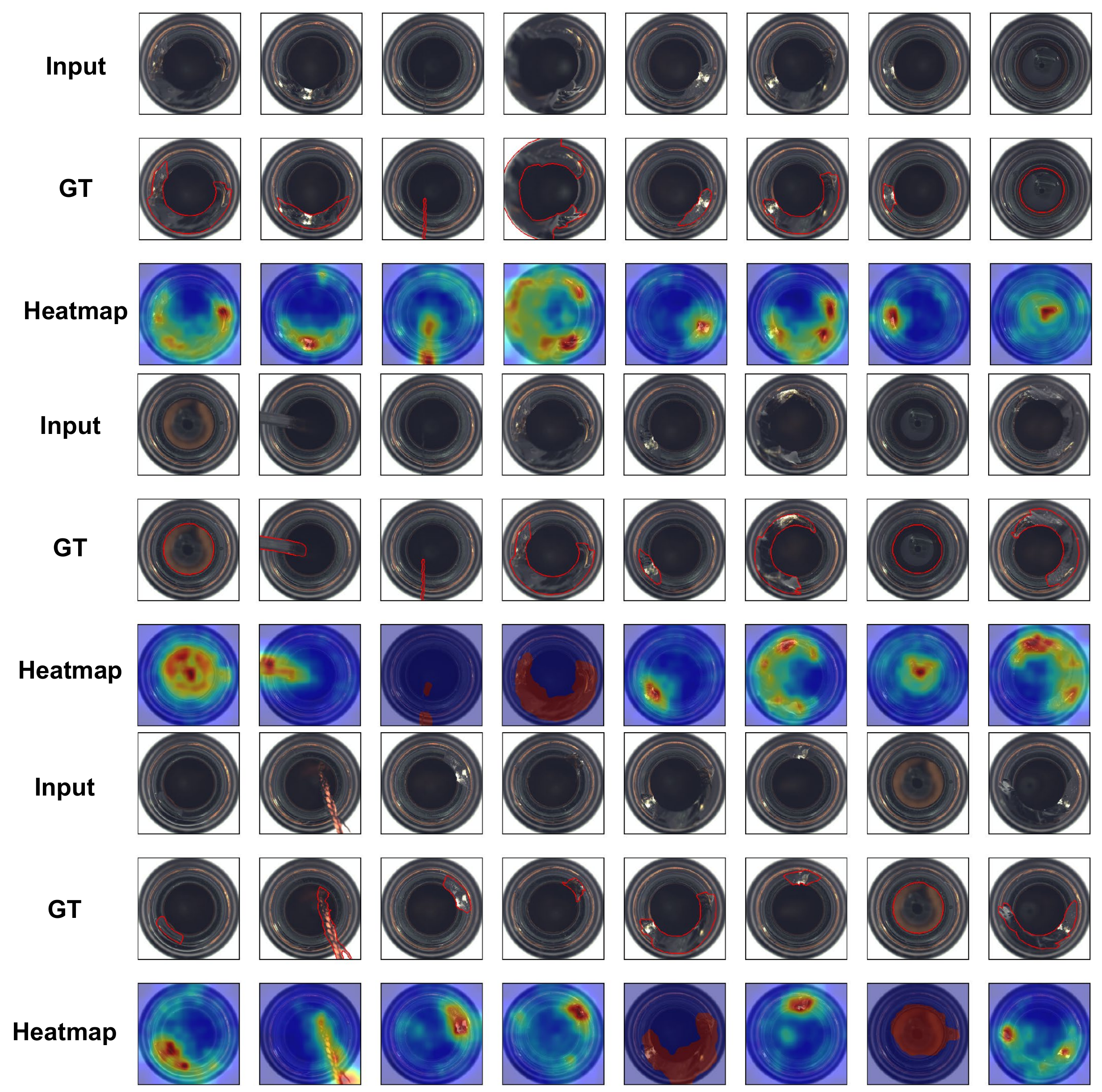}
  \caption{Anomaly localization on bottle class of MVTec AD. From top to bottom, input images, those with ground-truth localization area in red, heatmaps predicted by our model.}
  \label{fig:bottle}
\end{figure*}

\begin{figure*}[thp]
  \centering
  \includegraphics[width=\textwidth]{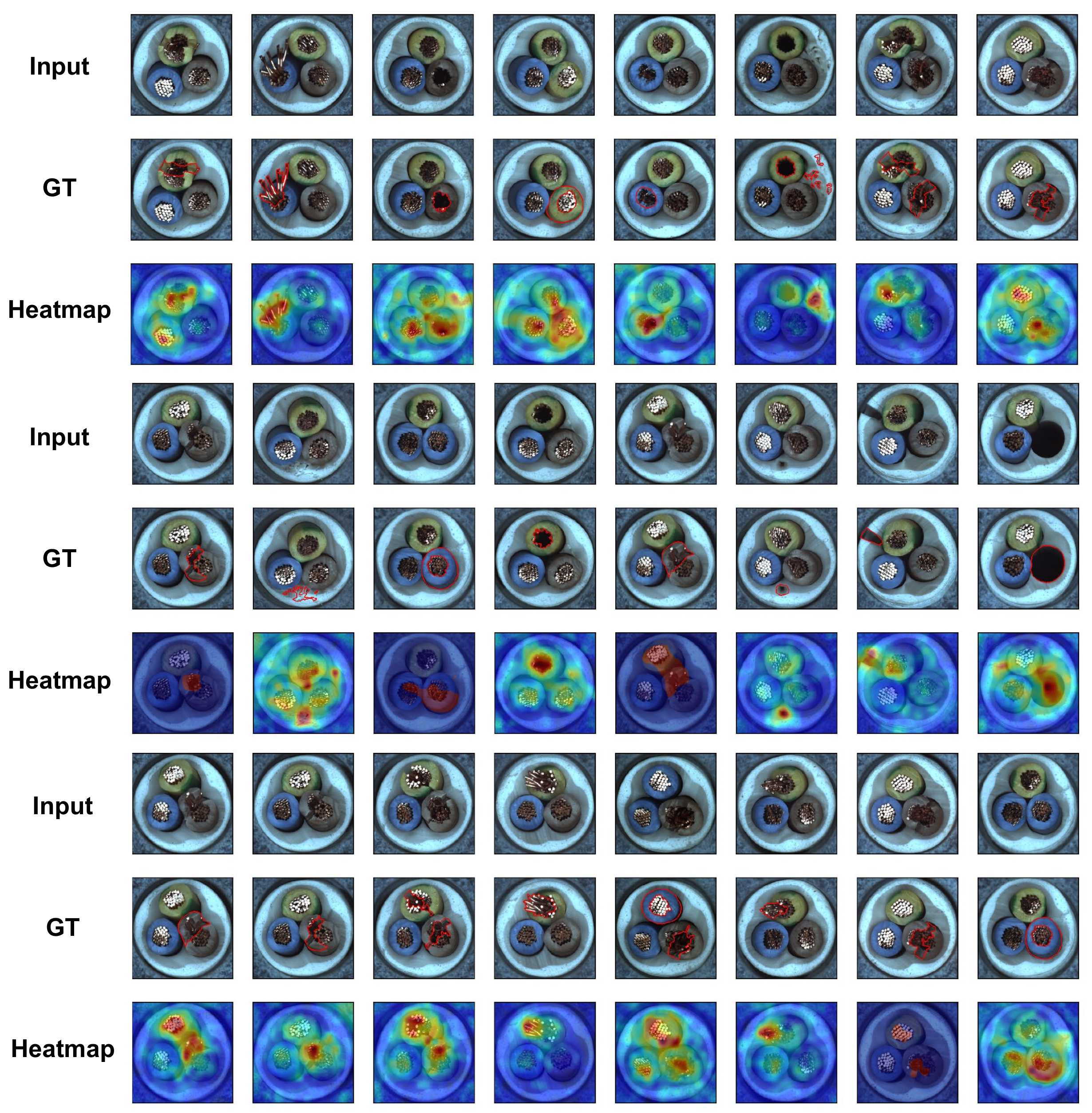}
  \caption{Anomaly localization on cable class of MVTec AD. From top to bottom, input images, those with ground-truth localization area in red, heatmaps predicted by our model.}
  \label{fig:cable}
\end{figure*}

\begin{figure*}[thp]
  \centering
  \includegraphics[width=\textwidth]{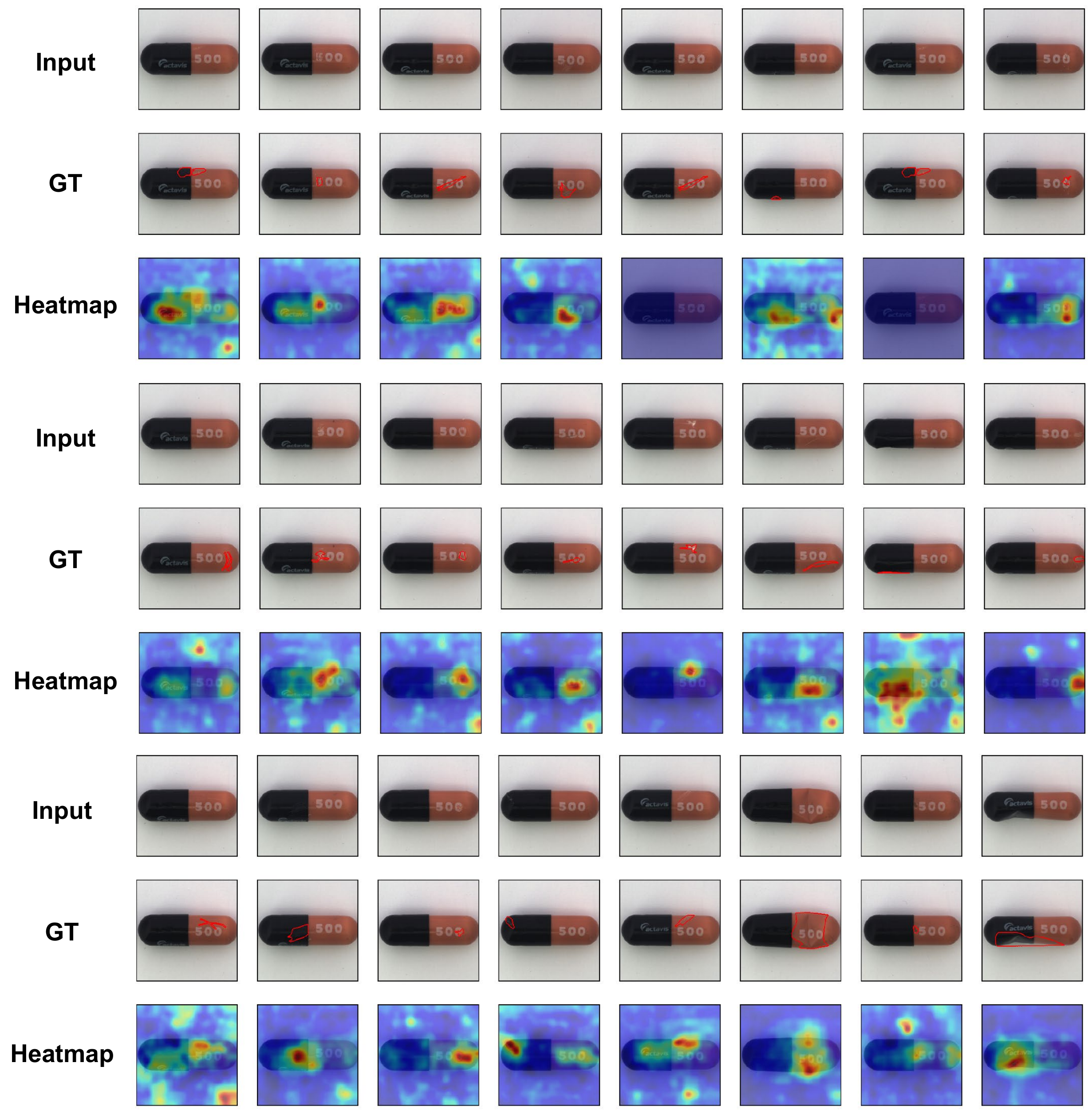}
  \caption{Anomaly localization on capsule class of MVTec AD. From top to bottom, input images, those with ground-truth localization area in red, heatmaps predicted by our model.}
  \label{fig:capsule}
\end{figure*}

\begin{figure*}[thp]
  \centering
  \includegraphics[width=\textwidth]{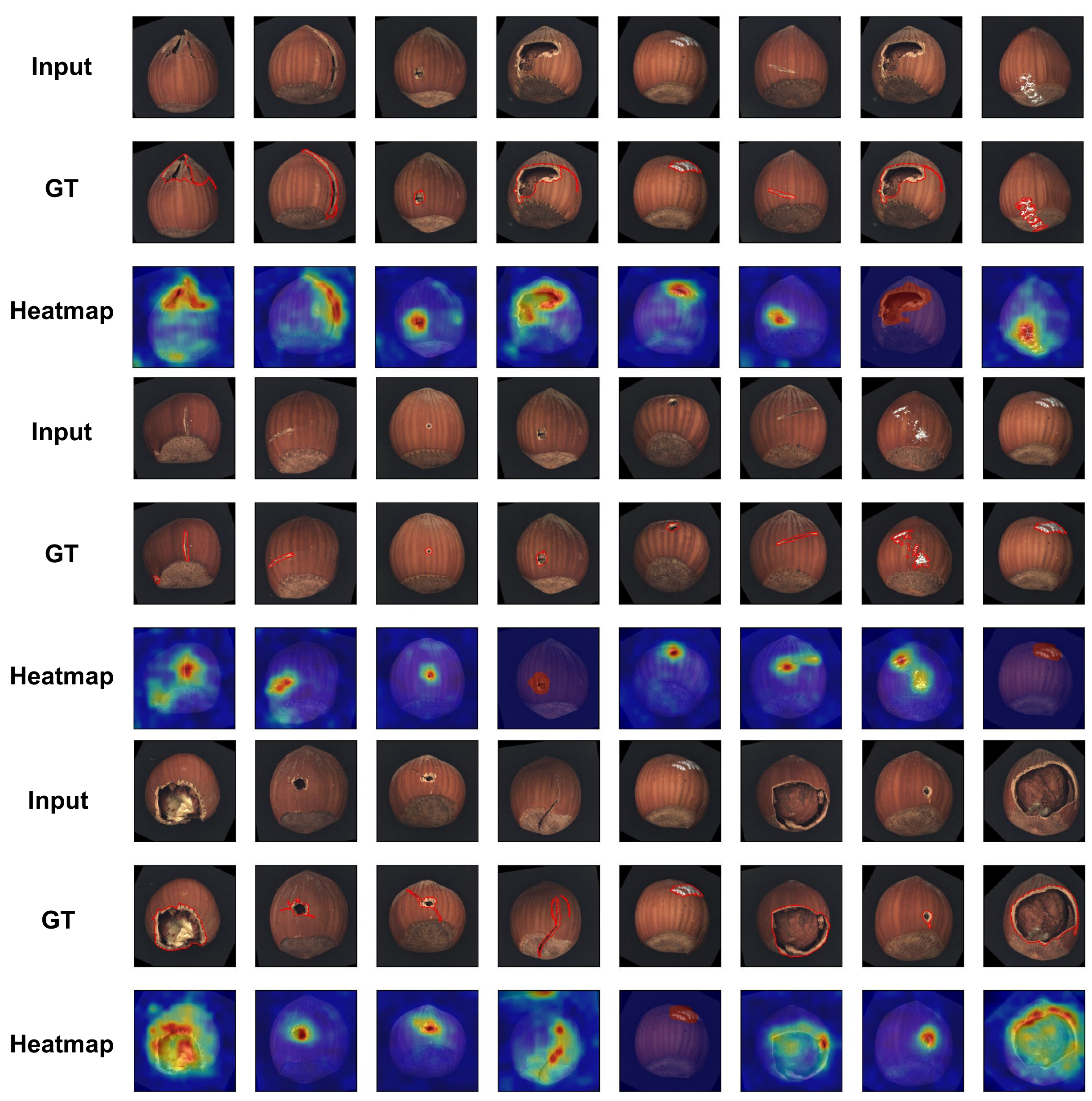}
  \caption{Anomaly localization on hazelnut class of MVTec AD. From top to bottom, input images, those with ground-truth localization area in red, heatmaps predicted by our model.}
  \label{fig:hazelnut}
\end{figure*}

\begin{figure*}[thp]
  \centering
  \includegraphics[width=\textwidth]{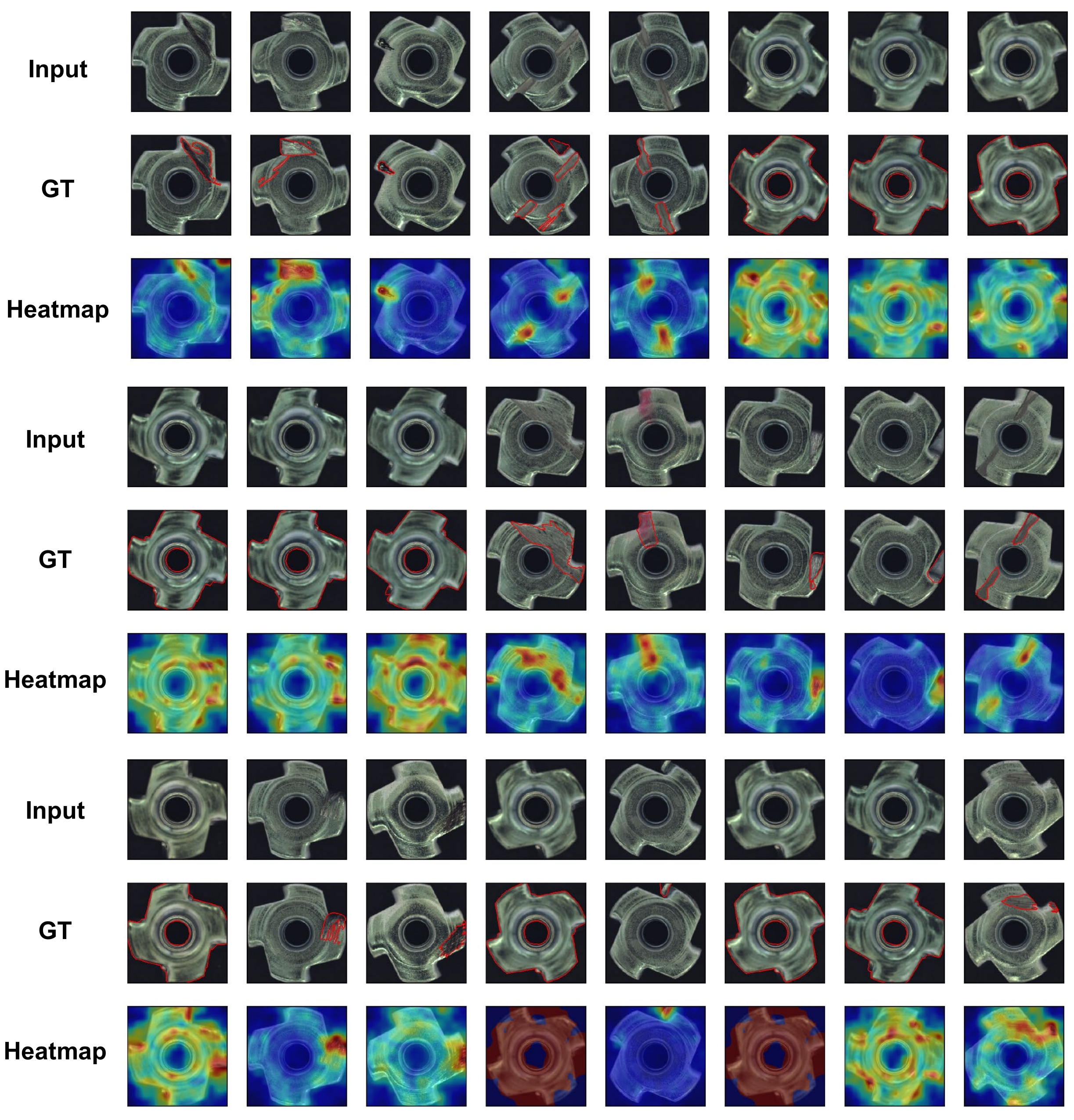}
  \caption{Anomaly localization on metal nut class of MVTec AD. From top to bottom, input images, those with ground-truth localization area in red, heatmaps predicted by our model.}
  \label{fig:metal_nut}
\end{figure*}

\begin{figure*}[thp]
  \centering
  \includegraphics[width=\textwidth]{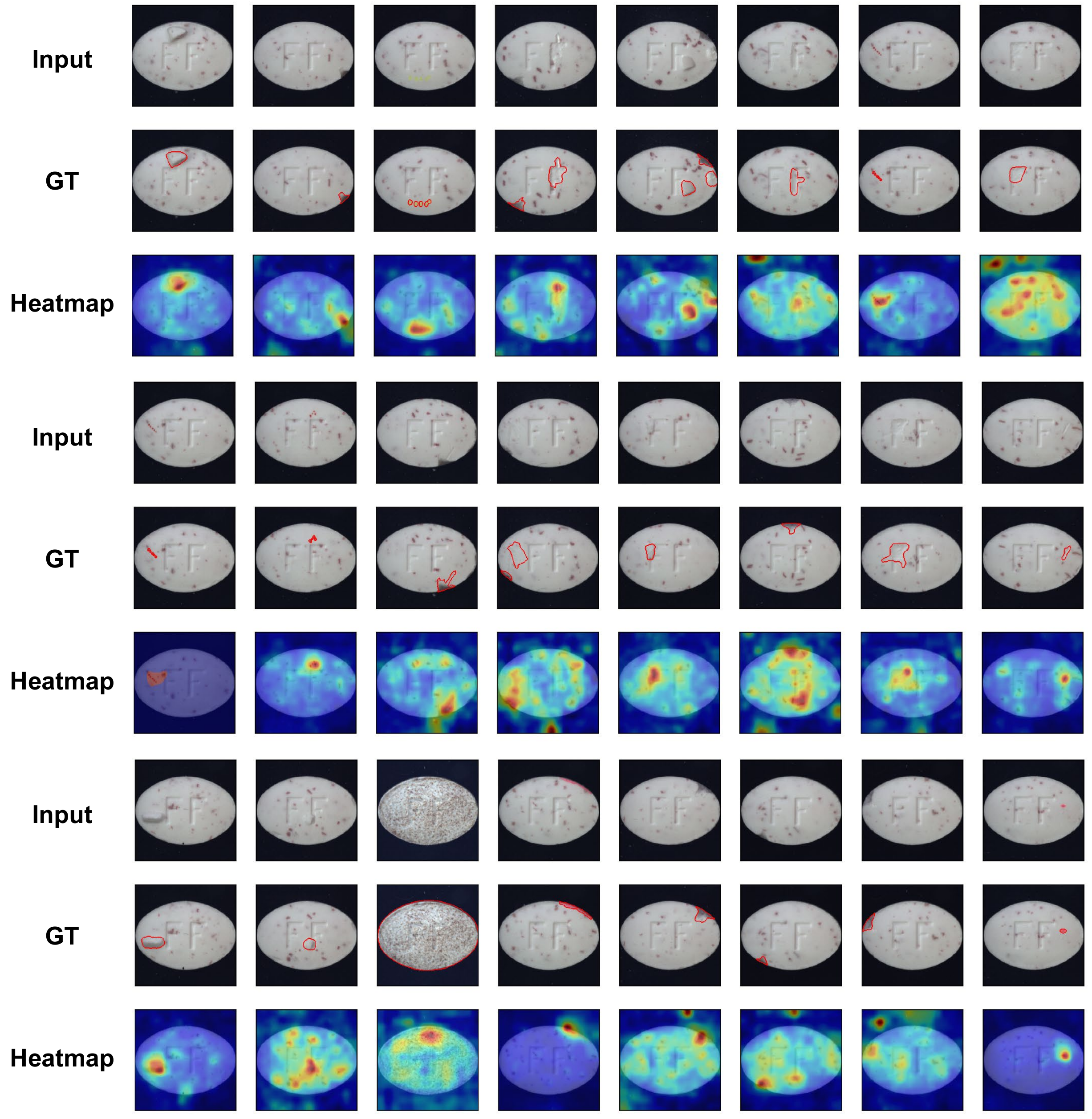}
  \caption{Anomaly localization on pill class of MVTec AD. From top to bottom, input images, those with ground-truth localization area in red, heatmaps predicted by our model.}
  \label{fig:pill}
\end{figure*}

\begin{figure*}[thp]
  \centering
  \includegraphics[width=\textwidth]{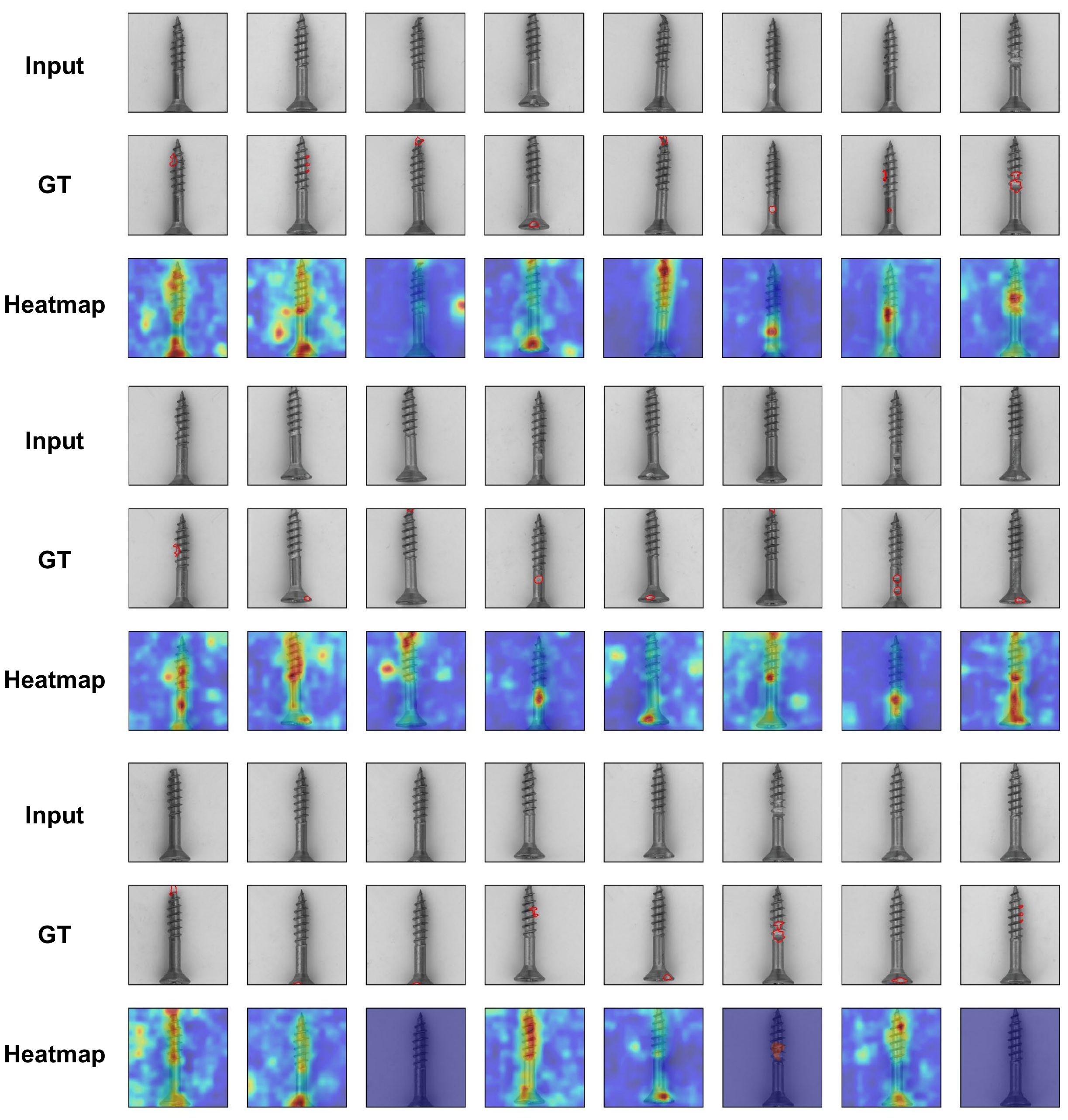}
  \caption{Anomaly localization on screw class of MVTec AD. From top to bottom, input images, those with ground-truth localization area in red, heatmaps predicted by our model.}
  \label{fig:screw}
\end{figure*}

\begin{figure*}[thp]
  \centering
  \includegraphics[width=\textwidth]{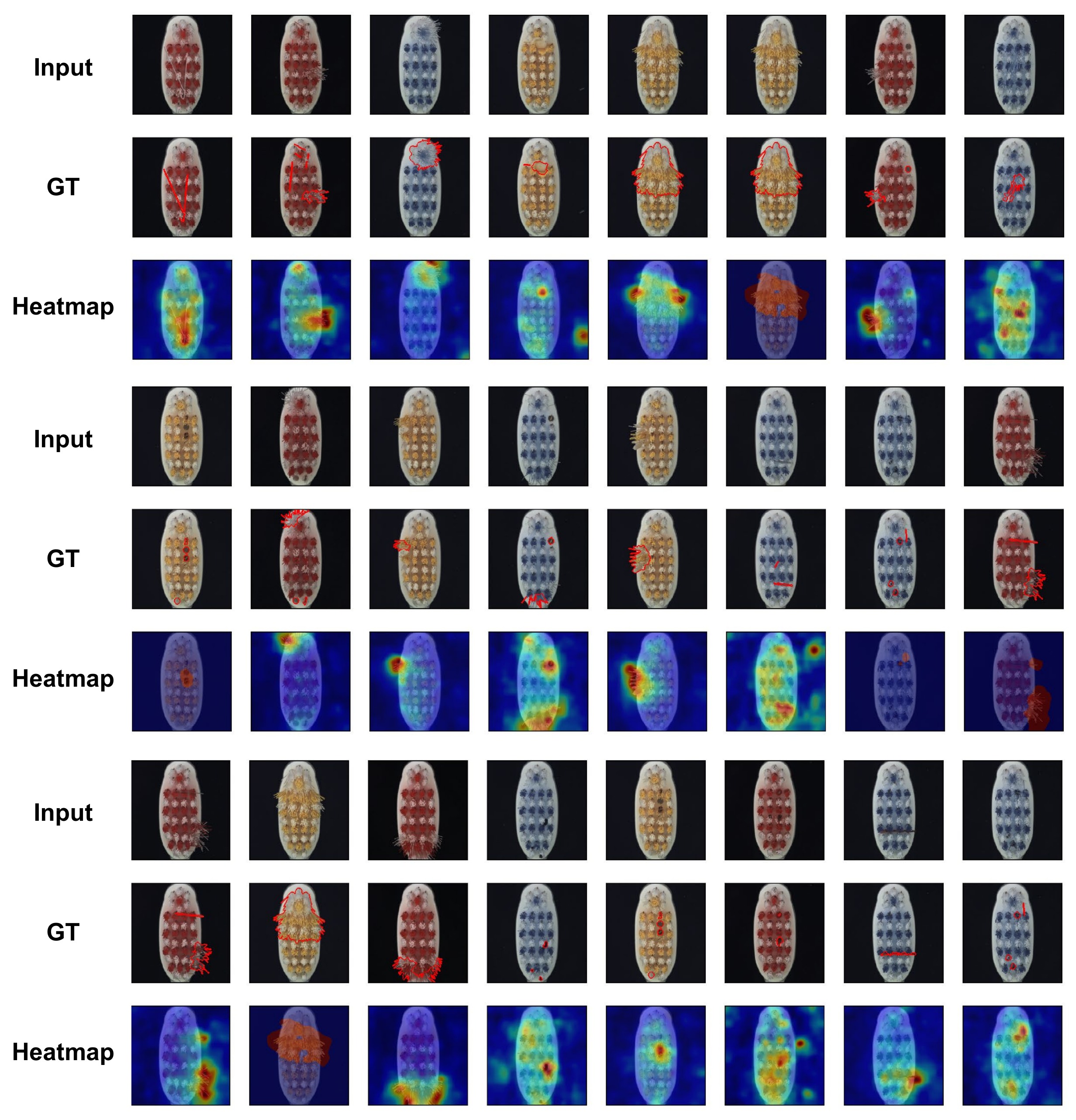}
  \caption{Anomaly localization on toothbrush class of MVTec AD. From top to bottom, input images, those with ground-truth localization area in red, heatmaps predicted by our model.}
  \label{fig:toothbrush}
\end{figure*}

\begin{figure*}[thp]
  \centering
  \includegraphics[width=\textwidth]{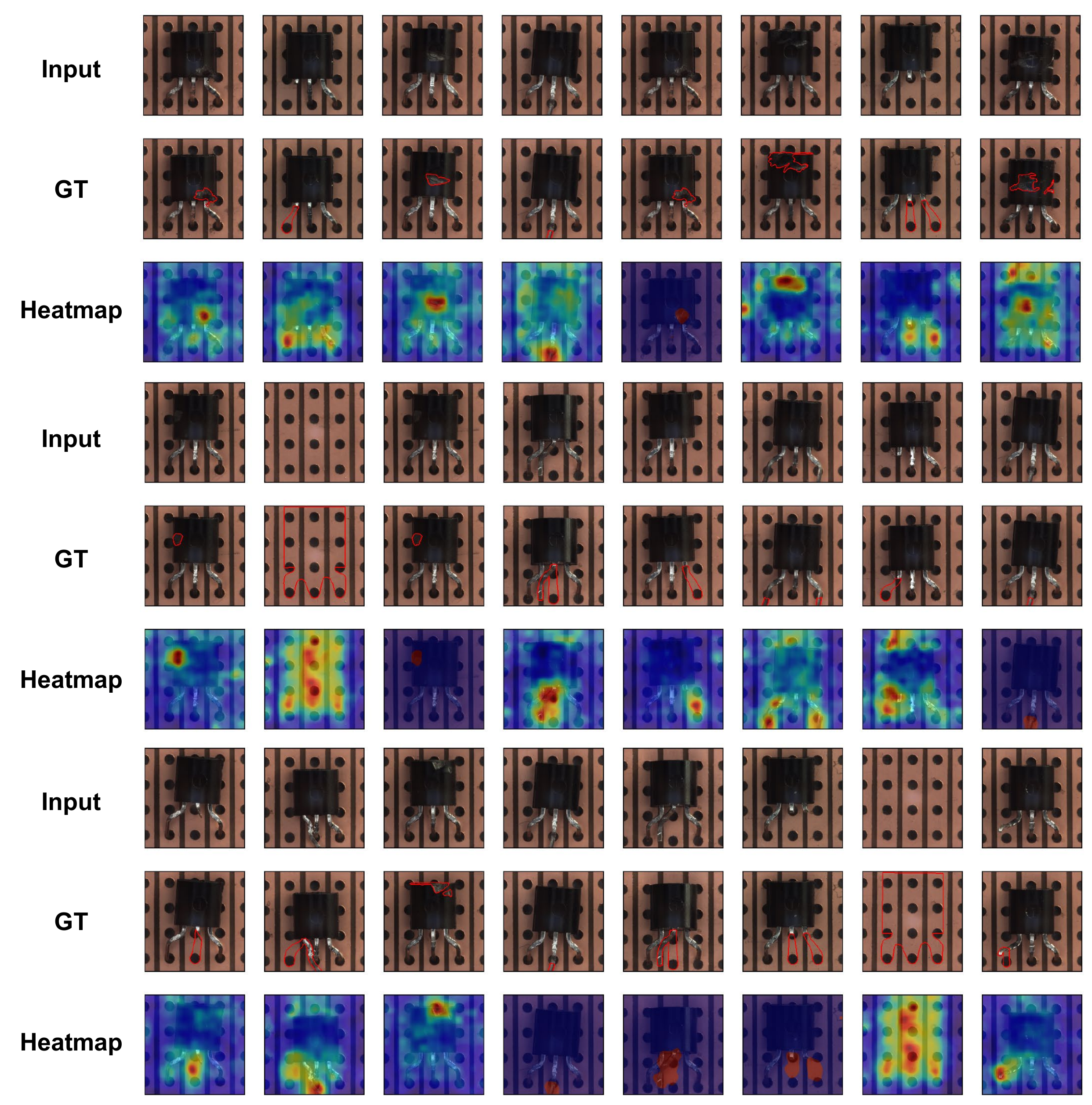}
  \caption{Anomaly localization on transistor class of MVTec AD. From top to bottom, input images, those with ground-truth localization area in red, heatmaps predicted by our model.}
  \label{fig:transistor}
\end{figure*}

\begin{figure*}[thp]
  \centering
  \includegraphics[width=\textwidth]{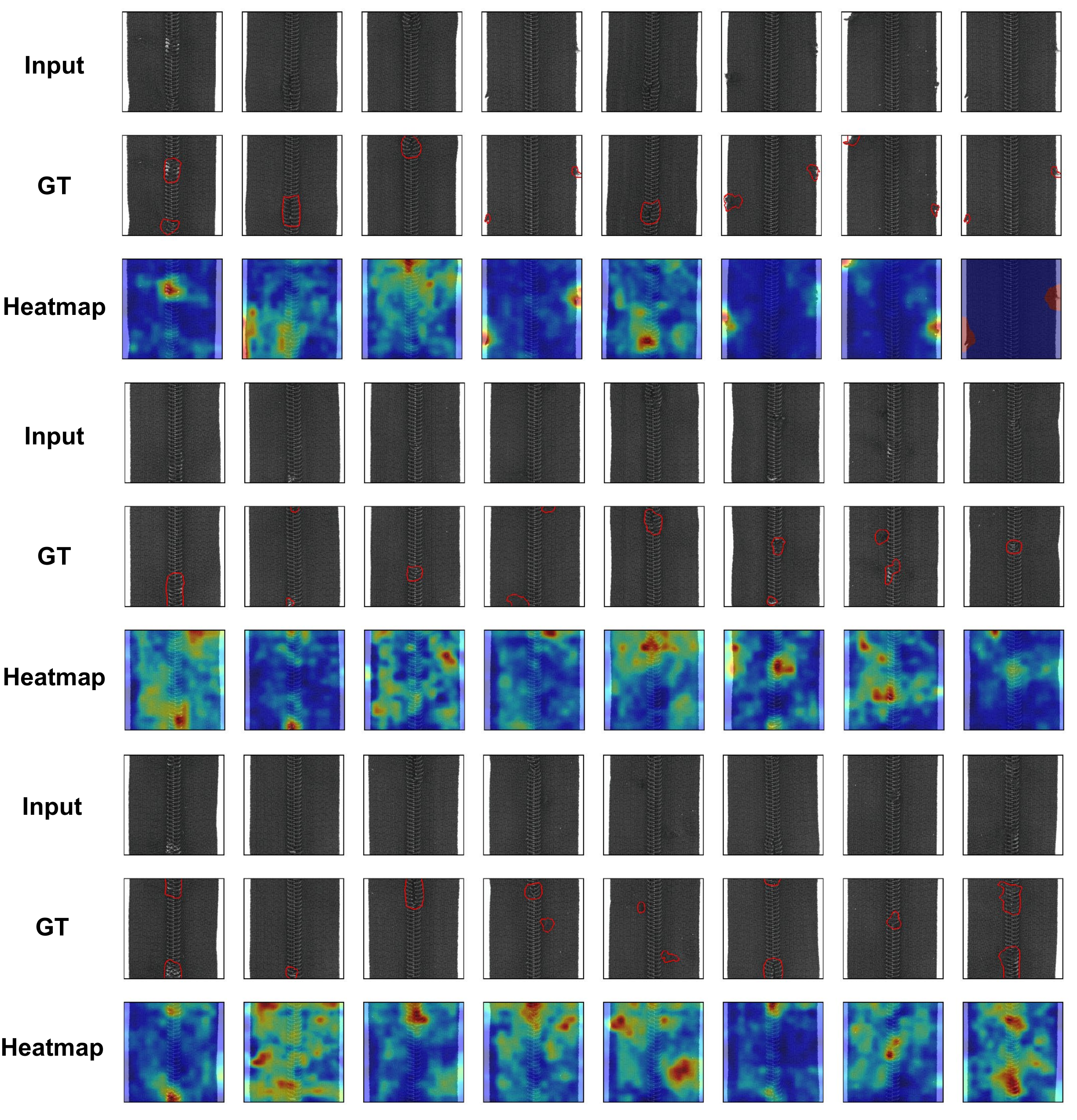}
  \caption{Anomaly localization on zipper class of MVTec AD. From top to bottom, input images, those with ground-truth localization area in red, heatmaps predicted by our model.}
  \label{fig:zipper}
\end{figure*}

\begin{figure*}[thp]
  \centering
  \includegraphics[width=\textwidth]{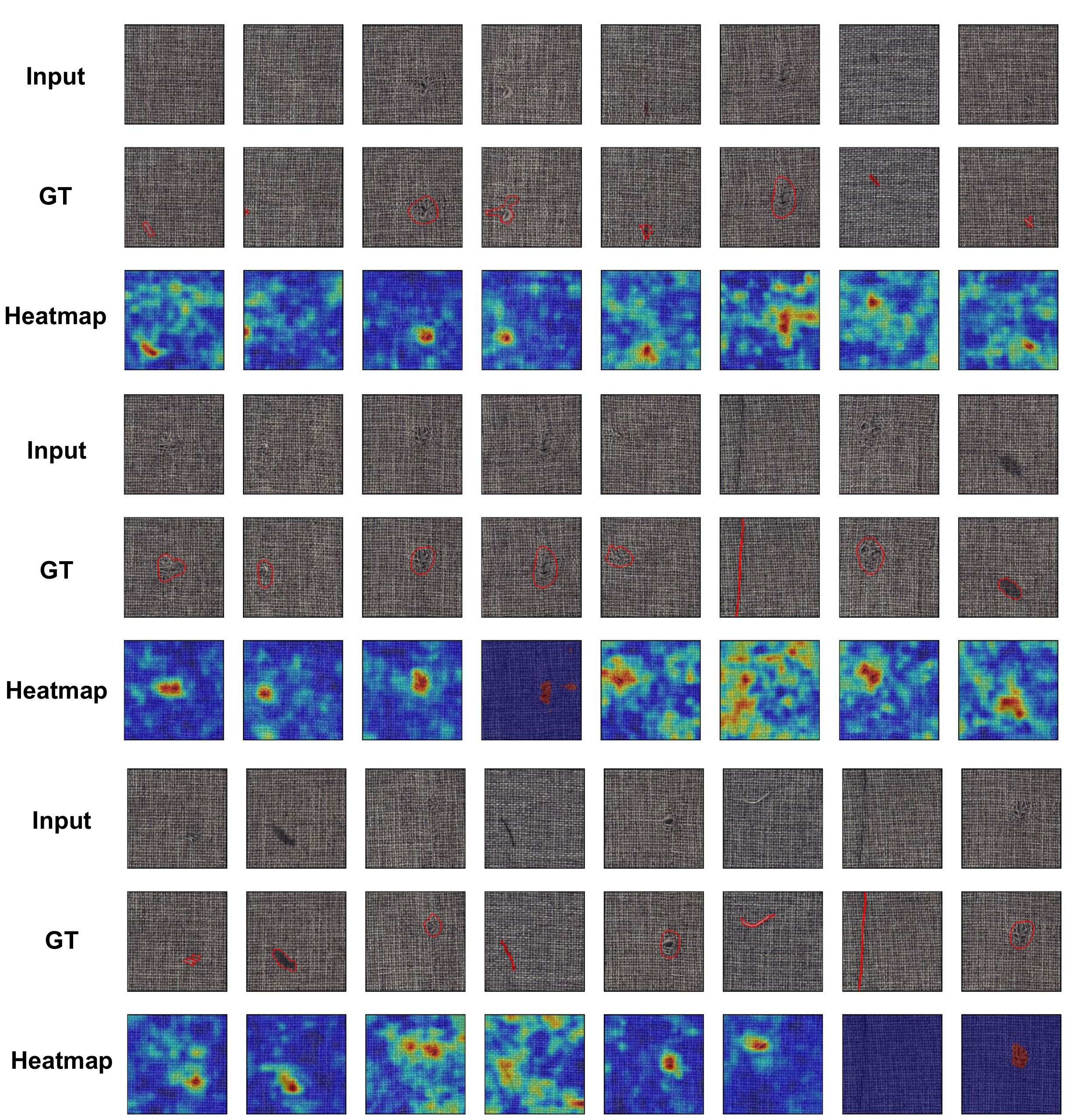}
  \caption{Anomaly localization on carpet class of MVTec AD. From top to bottom, input images, those with ground-truth localization area in red, heatmaps predicted by our model.}
  \label{fig:carpet}
\end{figure*}

\begin{figure*}[thp]
  \centering
  \includegraphics[width=\textwidth]{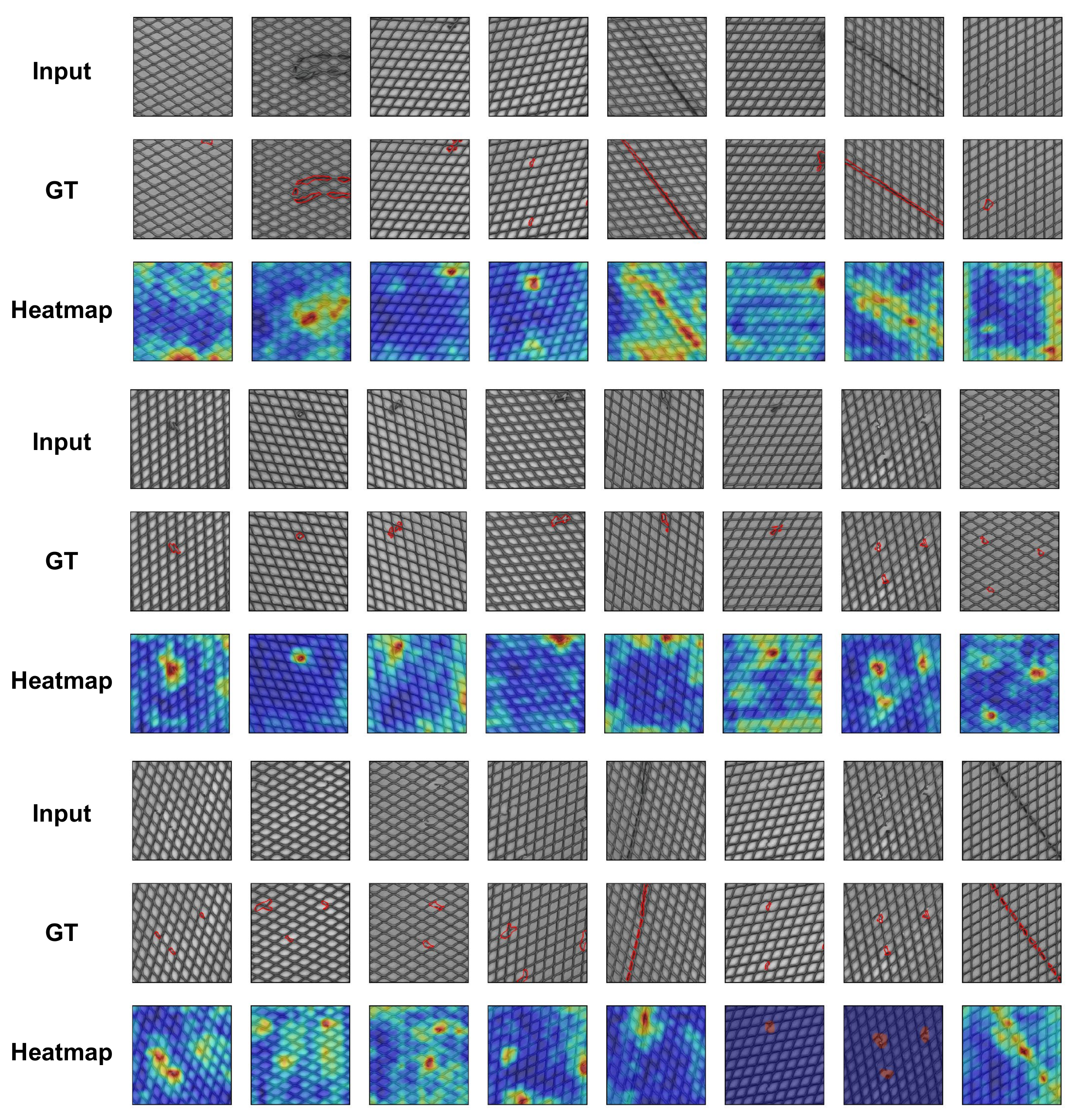}
  \caption{Anomaly localization on grid class of MVTec AD. From top to bottom, input images, those with ground-truth localization area in red, heatmaps predicted by our model.}
  \label{fig:grid}
\end{figure*}

\begin{figure*}[thp]
  \centering
  \includegraphics[width=\textwidth]{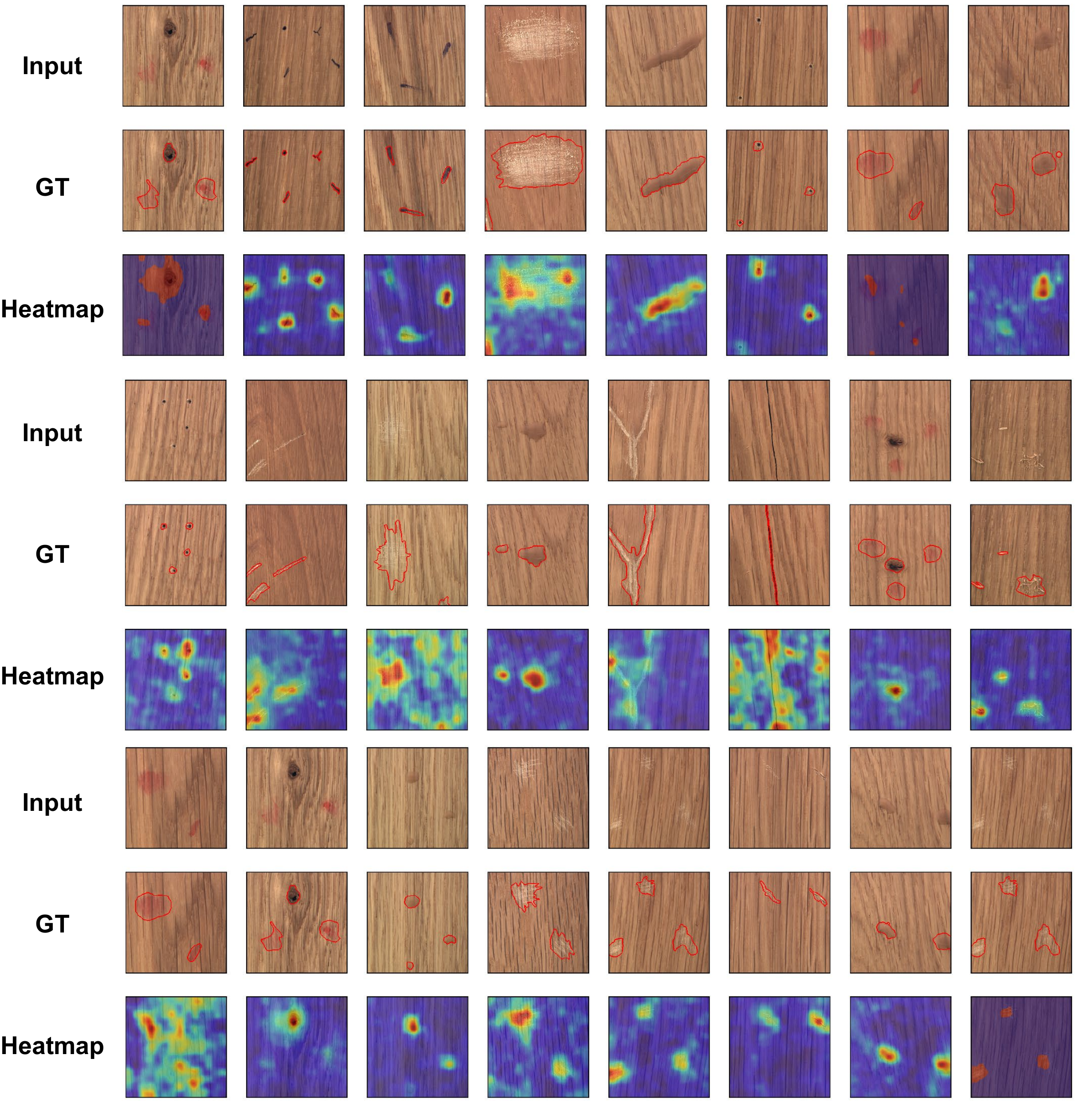}
  \caption{Anomaly localization on wood class of MVTec AD. From top to bottom, input images, those with ground-truth localization area in red, heatmaps predicted by our model.}
  \label{fig:wood}
\end{figure*}

\begin{figure*}[thp]
  \centering
  \includegraphics[width=\textwidth]{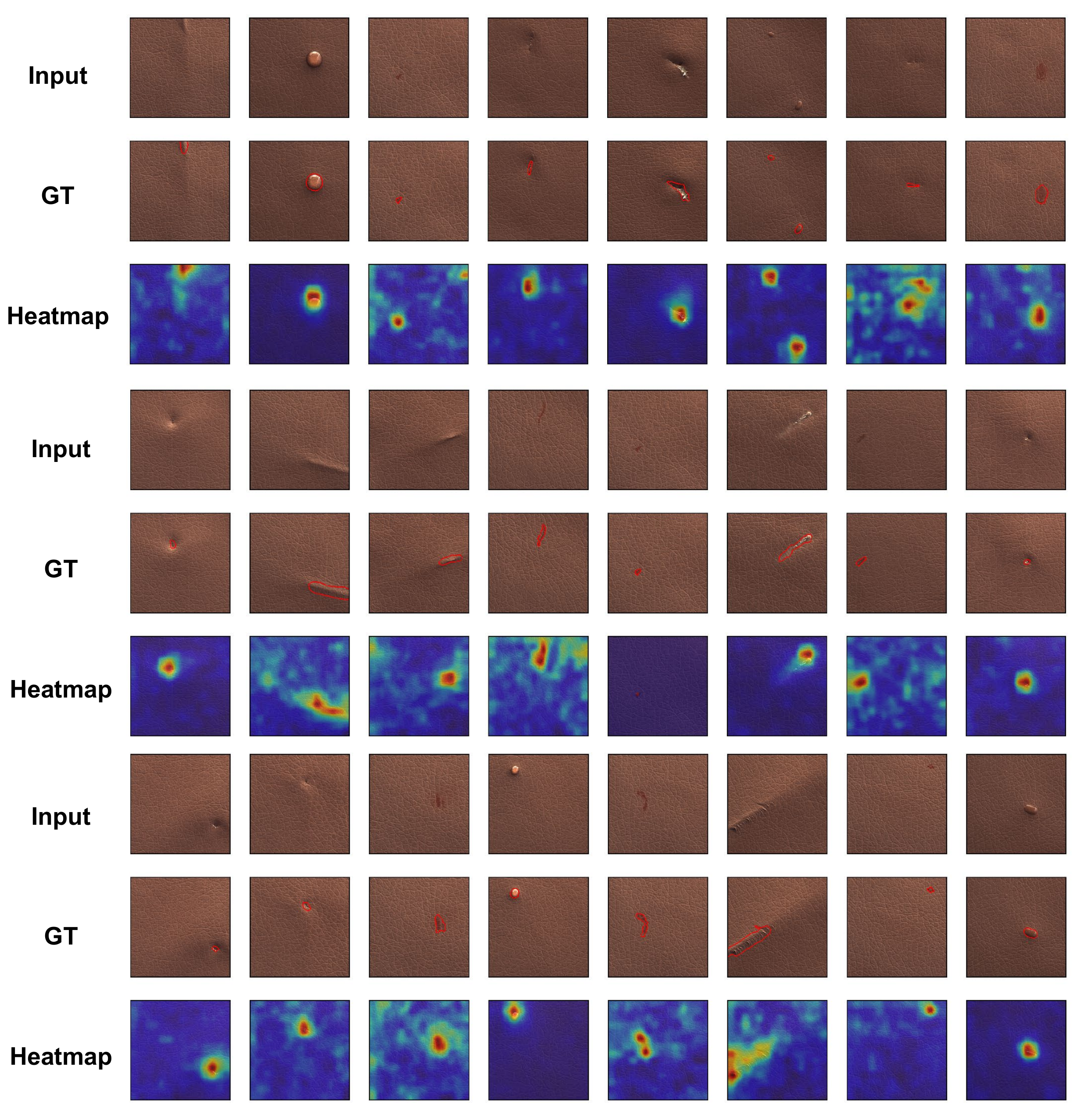}
  \caption{Anomaly localization on leather class of MVTec AD. From top to bottom, input images, those with ground-truth localization area in red, heatmaps predicted by our model.}
  \label{fig:leather}
\end{figure*}

\begin{figure*}[thp]
  \centering
  \includegraphics[width=\textwidth]{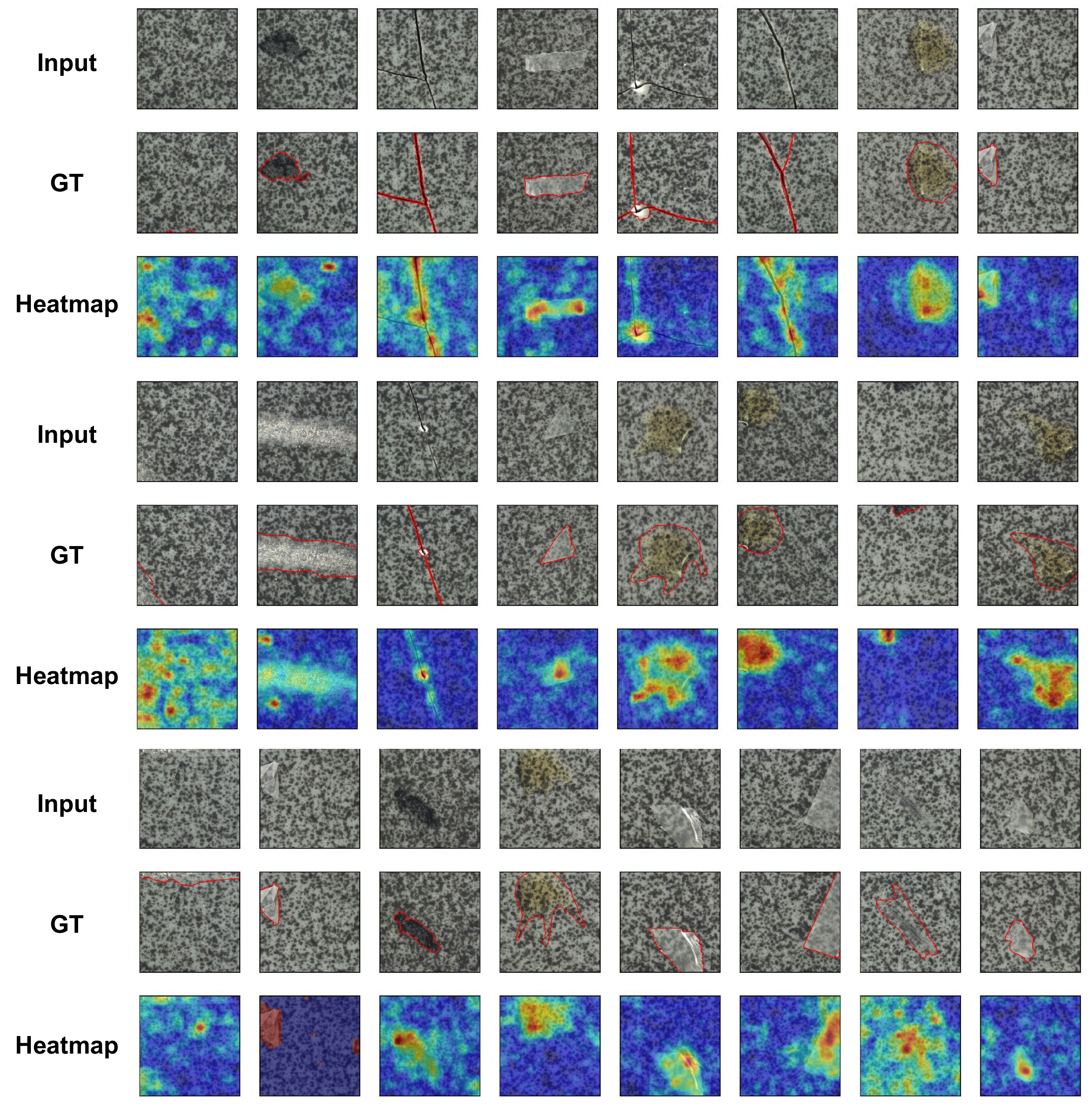}
  \caption{Anomaly localization on tile class of MVTec AD. From top to bottom, input images, those with ground-truth localization area in red, heatmaps predicted by our model.}
  \label{fig:tile}
\end{figure*}

From Figure~\ref{fig:bottle} to Figure~\ref{fig:tile}, we show localization visualizations examples of 10 object and 5 texture categories. We not only show successful cases, but also some failure cases.

\clearpage
\bibliography{aaai22}
\end{document}